\documentclass{article}

\usepackage[nonatbib,final]{neurips_2024}

\usepackage[utf8]{inputenc} 
\usepackage[T1]{fontenc}    
\usepackage{url}            
\usepackage{booktabs}       
\usepackage{amsfonts}       
\usepackage{nicefrac}       
\usepackage{microtype}      
\usepackage{xcolor}         
\usepackage{graphicx}
\usepackage{amsmath}
\usepackage{amssymb}
\usepackage{booktabs}
\usepackage{bbm}
\usepackage{amsthm}
\usepackage{xspace}
\usepackage{subfig}
\usepackage{booktabs}
\usepackage{multirow}
\usepackage{makecell}
\usepackage{color}

\usepackage{epsfig}
\usepackage[title]{appendix}
\usepackage{amsfonts}

\usepackage{graphicx}
\usepackage{amssymb}
\usepackage{amsthm}
\usepackage{mathrsfs}
\usepackage{algorithm}
\usepackage{algpseudocode}
\usepackage{amsfonts}
\usepackage{soul}
\usepackage{float}
\floatname{algorithm}{Protocol}
\usepackage{xspace}
\newcommand{\sysname}{TPA\xspace}

\newtheorem{thm}{\bf Theorem}[section]
\newtheorem{assumption}{Assumption}[section]

\usepackage[numbers,sort&compress]{natbib}
\usepackage[utf8]{inputenc} 
\usepackage[T1]{fontenc}    
\usepackage{hyperref}       
\usepackage{url}            
\usepackage{booktabs}       
\usepackage{amsfonts}       
\usepackage{nicefrac}       
\usepackage{microtype}      
\usepackage{xcolor}         
\usepackage{wrapfig}

\title{Transferability Bound Theory: Exploring Relationship between Adversarial Transferability and Flatness}

\author{%
	Mingyuan Fan\textsuperscript{1}, 
        Xiaodan Li\textsuperscript{1,2}, 
        Cen Chen\textsuperscript{1}\thanks{Corresponding author.}, 
        Wenmeng Zhou\textsuperscript{2}, 
        Yaliang Li\textsuperscript{3} \\
	\textsuperscript{1}School of Data Science \& Engineering, East China Normal University, China \\
	\textsuperscript{2}Alibaba Group, Hangzhou, China \\
	\textsuperscript{3}Alibaba Group, Bellevue, WA, USA \\
	\texttt{fmy2660966@gmail.com, cenchen@dase.ecnu.edu.cn}, \\
	\texttt{\{fiona.lxd, wenmeng.zwm, yaliang.li\}@alibaba-inc.com}
}

\begin{document}

\maketitle

\begin{abstract}
    A prevailing belief in attack and defense community is that the higher flatness of adversarial examples enables their better cross-model transferability, leading to a growing interest in employing sharpness-aware minimization and its variants.
    However, the theoretical relationship between the transferability of adversarial examples and their flatness has not been well established, making the belief questionable.
    To bridge this gap, we embark on a theoretical investigation and, for the first time, derive a theoretical bound for the transferability of adversarial examples with few practical assumptions.
    Our analysis challenges this belief by demonstrating that the increased flatness of adversarial examples does not necessarily guarantee improved transferability.
    Moreover, building upon the theoretical analysis, we propose \sysname, a Theoretically Provable Attack that optimizes a surrogate of the derived bound to craft adversarial examples.
    Extensive experiments across widely used benchmark datasets and various real-world applications show that \sysname can craft more transferable adversarial examples compared to state-of-the-art baselines.
    We hope that these results can recalibrate preconceived impressions within the community and facilitate the development of stronger adversarial attack and defense mechanisms.
    The source codes are available in \url{https://github.com/fmy266/TPA}.
\end{abstract}

\maketitle

\section{Introduction}
\label{intro}
The transferability of adversarial examples \cite{TI,trust_survey} suggests a phenomenon where adversarial examples designed to fool one local proxy model can also be generalized to mislead other unknown target models, even with different model parameters and architectures.
This intriguing property has significant implications for various applications, ranging from the robustness assessment of large-scale models~\cite{split_adv,SSA} to data privacy protection~\cite{unlearnable_samples,adv_1,adv_2}.
However, it is observed that the generated adversarial examples often tend to be overly specialized to the proxy model, showing limited transferability \cite{VR,SSA,RAP}.
Driven by this, researchers have dedicated substantial efforts to developing transferability-enhancing techniques \cite{RAP,styless,BSR}.

Most transferability-enhancing techniques have drawn inspiration from an analogy between the transferability of adversarial examples and the generalization capability of models \cite{DI,NI_SI,VR}.
For example, borrowing the idea of Mixup~\cite{mixup}, Admix \cite{admix} attempts to blend adversarial examples with other samples to generate more transferable adversarial examples.
Recently, \citet{SAM4} observed a strong correlation between the flatness of the model's loss landscape and its generalization capability, thus advocating for the optimization of the worst-case loss within a small perturbation radius to regulate the loss landscape.
The empirical performance of such sharpness-aware minimization is quite impressive, sparking a surge of interest among researchers and catalyzing the development of its variants.
Given the above considerable progress, the exploitation of sharpness-aware minimization and its variants to enhance the transferability of adversarial examples has become one of the hottest research topics \cite{flatattack1,flatattack2,flatattack3,flatattack4}.

Although empirical results show that transferability-enhancing techniques based on flatness achieve state-of-the-art performance~\cite{RAP,flatattack1,flatattack2}, the theoretical relationship between adversarial transferability and flatness has not been well established.
As such, it remains questionable whether adversarial examples converging to flat extreme points, referred to as \textit{flat adversarial examples}, necessarily possess better transferability.
In fact, the relationship between flatness and generalization capability is still under debate and not well-solved.
Recent works \cite{SAM7,SAM8,SAM9} have brought to light a thought-provoking finding, i.e., flat extreme points do not always lead to better generalization ability, making the relationship between adversarial transferability and flatness even more mysterious.
Given the aforementioned gap in our current understanding, we first theoretically investigate the following critical question:

\begin{center}
\textbf{Q1:} \textit{Are flat adversarial examples necessarily associated with improved transferability?}
\end{center}

Our theoretical examination yields a negative answer to Q1.
Specifically, we derive a bound of adversarial transferability and demonstrate that, despite being a contributing term, flatness alone is insufficient to determine the value of the bound.
Then, our attention naturally shifts to explore Q2:
\begin{center}
    \textbf{Q2:} \textit{Can our theoretical analysis guide to develop a more principled transfer-based attack?}
\end{center}

One straightforward idea for addressing the Q2 is to directly optimize the derived bound.
However, the optimization of this bound requires higher-order gradient information, which is prohibitively expensive in high-dimensional spaces.
To circumvent this issue, we introduce Theoretically Provable Attack (\sysname), which employs a theoretically-grounded surrogate of the original bound as an alternative.
The optimization of the surrogate only requires first-order gradient information, rendering \sysname more computationally efficient and practical.

It should be stressed that the surrogate in \sysname is theoretically supported, ensuring that its attack effectiveness enjoys theoretical guarantees.
Moreover, the empirical evaluation of \sysname in the benchmarks and three kinds of real-world applications demonstrates the superior performance of \sysname over the state-of-the-art attack methods.
Our contributions are three-fold:
\begin{itemize}
    \item To our best knowledge, we are the first to derive a theoretical bound on the transferability of adversarial examples, with few practical assumptions.
    Contrary to the prevalent belief among the community, we demonstrate that the flatness of adversarial examples does not necessarily guarantee enhanced transferability.
    Our theory also advances the understanding of the intrinsic characteristics governing transferability and lays a solid theoretical foundation for future research in this area.
    \item Building upon our theoretical analysis, we introduce \sysname which optimizes a surrogate of the original bound to avoid direct computation of high-order gradients.
    We demonstrate the effectiveness of the surrogate as the surrogate of the derived bound.
    \item We conduct extensive experiments in the benchmark dataset ImageNet, showing the impressive performance of \sysname. \textit{To our best knowledge, \sysname is the first transfer-based attack to achieve an average 90\% attack success rate against transformer architectures.} We also test the effectiveness of \sysname in three kinds of real-world applications that are under-explored in the existing works, including Google Vision Systems, advanced search engines, as well as multimodal large model applications (e.g., GPT-4 and Claude3).
\end{itemize}

\section{Related Work}
\label{related_works}
\subsection{Flatness-based Optimization Methods}

Intuitively, the loss landscape of a model on training data mirrors that on test data, suggesting that a slight translation of the loss landscape over training data can approximate the loss landscape over test data.
Given this, a flat extreme point is believed to enjoy a smaller performance gap between training and testing data than sharp extreme points, i.e., better generalization ability.
Capitalizing on this intuition, SAM \cite{SAM2,SAM3,SAM4} incorporates the following flatness term or its variants into its loss function to improve the model's generalization capability:
\begin{equation}
\label{sam_term}
\max_{||\rho|| \leq \epsilon} \mathcal{L}(F_{\theta+\rho}(x),y) - \mathcal{L}(F_{\theta}(x),y),
\end{equation}
where $(x,y), \mathcal{L}, F_{\theta}$ and $\epsilon$ are training data pairs, loss function, model parameterized by $\theta$, and perturbation budget, respectively.
Equation \ref{sam_term} evaluates the loss landscape's flatness at point $\theta$ by quantifying the increase in loss when moving from $\theta$ to a neighboring worst-behaved parameter value.
Subsequent studies explored various flatness metrics, such as Fisher determinant \cite{SAM1} and gradient norms \cite{SAM5,SAM6}, for regularizing loss landscapes.
Nevertheless, recent works \cite{SAM7,SAM8,SAM9,SAM10} have cast a critical light.
\citet{SAM7} demonstrated the existence of networks that have a sharp loss landscape yet manage to work well on test data.
\citet{SAM8} empirically revealed that there is not a strong correlation between flatness and generalization ability, particularly in large neural networks.

\subsection{Transferability-enhancing Methods}

Transferability-enhancing methods can be roughly divided into input-regularization-based methods, optimization-based methods, and model-based methods.
In each iteration, input-regularization-based methods \citep{pretrainedadv,DI,VR,admix} ensemble multiple transformed inputs to craft adversarial examples, and the distinction between these methods is reflected in the adopted transformation techniques.
DI \citep{DI} suggests to resize and pad inputs, while TI \citep{TI} translates the inputs and SI \citep{NI_SI} scales the inputs.
By observing that existing transformations are all applied on a single image, Admix \citep{admix} attempts to admix inputs with an image from other categories.
Unlike the spatial domain transformations mentioned above, SSA \citep{SSA} perturbs inputs in frequency domains to produce more diverse transformed inputs.
BSR \citep{BSR} segments inputs into non-overlapping blocks, shuffling and rotating the blocks randomly.

Optimization-based methods leverage more advanced optimizers \citep{NI_SI,MI} and model-based methods enhance the transferability via the lens of model per se \citep{ghost_net,dra}.
For instance, MI \citep{MI} adopts Momentum optimizer.
Besides, SGM \citep{sgm} refines the back-propagation procedure to amplify the gradients of early layers, due to that the features learned by early layers are more shared over different models.
StyLess \citep{styless} employs stylized networks to prevent adversarial examples from using non-robust style features.
More recent advancements delved into the exploration of flatness-based optimization methods to improve the transferability of adversarial examples.
\citet{RAP} introduced RAP, which tailors Equation \ref{sam_term} to craft flatness-enhanced adversarial examples.
Some studies \cite{flatattack1,flatattack2} chose to punish the gradients of the generated adversarial examples.


\section{Theoretical Analysis}
\label{sec_theory}

In this section, we embark on a theoretical exploration of adversarial transferability, commencing with the canonical setup of transfer-based adversarial attacks.
The primary objective of transfer-based attacks is to make adversarial examples crafted using the local proxy model $F$ as effective as possible for the target model $F'$.
Formally, for a given natural samples $x$ with ground-truth labels $y$, the vanilla transfer-based attack solves the following optimization task to craft adversarial noises $\delta^*$ for $x$ as:
\begin{equation}
\label{vanilla_optim}
\delta^* = \underset{\delta}{\arg \min}  -\mathcal{L}(F(x+\delta),y), \quad s.t., \ 
||\delta||_{\infty} \leq \epsilon,
\end{equation}
where $\mathcal{L}$ is cross-entropy loss and $\epsilon$ is perturbation budget.
Despite the high attack effectiveness of $x+\delta^*$ on the proxy model, their effectiveness significantly diminishes when transferred to the target model.
To better study this problem, we decouple transferability into two factors: \textit{the local effectiveness term}, which measures the loss of generated adversarial examples on the proxy model, along with \textit{the transfer-related loss term}, which quantifies the change in the loss of adversarial examples when transferring from the proxy model to the target model.
The local effectiveness term and the transfer-related loss term can be evaluated by $\mathcal{L}(F(x+\delta),y)$ and $D(x+\delta,y)=\mathcal{L}(F'(x+\delta),y) - \mathcal{L}(F(x+\delta),y) $, respectively.
Intuitively, transferable adversarial examples are characterized by high attack performance on the proxy model together with minimal increase in the transfer-related loss term upon moving to the target model.
Before presenting our theoretical results, we first make several practical assumptions.

\begin{assumption}
\label{assumpt1}
The underlying distribution of $x$ is continuous and bounded: there exists a constant $B_1$ such that $\forall x$, it holds that $p(x) \leq B_1$.
\end{assumption}

\begin{assumption}
\label{assumpt2}
The gradient norm of $p(x)$ is bounded: there exists a constant $B_2$ such that $\forall x$, it holds that $||\nabla p(x)|| \leq B_2$.
\end{assumption}

\begin{assumption}
\label{assumpt3}
The probability of adversarial examples occurring in nature is less than that of natural samples: $p(x+\delta) \leq p(x)$.
\end{assumption}

\begin{assumption}
\label{assumpt4}
Proxy-model-generated adversarial examples have a greater loss on the proxy model than on the target model: $\mathcal{L}(F'(x+\delta),y) \leq \mathcal{L}(F(x+\delta),y)$.
\end{assumption}

\begin{assumption}
\label{assumpt5}
The proxy model is based on ResNet-like architecture, where each layer applies a linear transformation followed by a ReLU activation.
\end{assumption}

\begin{thm}
\label{thm1}
(See Appendix \ref{proof_theory} for Proof.) 
Let Assumption \ref{assumpt1} $\sim$ \ref{assumpt5} stand. For small $||\delta||_2^2$, it holds that:
\begin{equation}
\label{our_bound}
\begin{split}
&\mathbb{E}_{p(x)} \{||D(x+\delta,y) ||_2^2 \} \leq
\underbrace{\mathbb{E}_{p(x)} \{ ||D(x,y)||_2^2  + C \  ||\delta||_2^2 \ ||\nabla D(x,y)||_2^2 \}}_{\text{The inherent model difference component}} \\
&+ 
\underbrace{(1+C) \mathbb{E}_{p(x)} \{ ||\delta||_2^2 \ || \nabla \log F(x+\delta) ||_2^2 \} }_{\text{The first-order gradient component}} +
\underbrace{2 \ \mathbb{E}_{p(x)} \{ ||\delta||_2^2 \sum_{i} \ |\nabla^2 \log F(x+\delta)[i,i]| \}}_{\text{The second-order gradient component}},
\end{split}
\end{equation}
where $C=\frac{B_1}{B_1+B_2 ||\delta||}$\footnote{In fact, there is $C \leq 1$, and using $C$ of 1 could be useful in certain situations, such as calculating the specific value of upper bound for $||D(x+\delta,y)||_2^2$. Here, for the sake of the tightness of our bound, we have not done so.} and $[i,i]$ denotes the element at the intersection of the $i$-th row and the $j$-th column in the given matrix.
Let the sum of the three terms on the right side of Equation \ref{our_bound} be denoted as $K$.
Then, we have:
\begin{equation}
||\mathcal{L}(F'(x+\delta),y)||_2^2 \geq |(||\mathcal{L}(F(x+\delta),y)||_2^2 - K)|.
\end{equation}
\end{thm}

We use $p(x)$ to denote the underlying distribution of $x$ and posit Assumption \ref{assumpt1} $\sim$ \ref{assumpt5}.
Assumption \ref{assumpt1} arises naturally since natural samples do not occur with infinite probability.
Building upon Assumption \ref{assumpt1}, a straightforward deduction is that the gradients of $p(x)$ are bounded; otherwise, there exists at least one sample that appears with infinite probability.
Assumptions \ref{assumpt3} and Assumption \ref{assumpt4} are naturally valid.
Regarding Assumption \ref{assumpt3}, adversarial samples are specifically crafted rather than commonly occurring in nature.
Moreover, discussions about transfer-based attacks only make sense under the conditions set by Assumption \ref{assumpt4}.
Assumption \ref{assumpt5} is satisfiable because the attacker can freely choose which proxy model to use.
Theorem \ref{thm1} establishes bounds for $||D(x+\delta,y)||_2^2$ and $||\mathcal{L}(F'(x+\delta),y)||_2^2$.
We mainly discuss the bound of $||D(x+\delta,y)||_2^2$, which is composed of three key components, namely the inherent model difference, the first-order gradient, and the second-order gradient components.
For the sake of simplicity in our discussion, $||\delta||_2^2$ is considered as a constant.

\textbf{The inherent model difference component.}
This component is intuitively interpreted as a measure of the similarity between the proxy and target models by evaluating their output differences regarding natural samples $x$.
A higher value of the component implies that adversarial examples produced by the proxy model are less likely to transfer to the target model.

In practice, the value of this component tends to be modest due to the following reasons.
Firstly, the target model is expected to perform well on $x$, as misclassification of the target model to $x$ would render crafting adversarial examples for $x$ trivial.
Secondly, as the attacker is capable of fine-tuning the proxy model on $x$, the loss of $x$ in the proxy model can also be small.
Furthermore, if the discrepancy between the two models is small, their gradient differences also are minor as well~\cite{split_adv_attack}.
Thus, both $||D(x,y)||_2^2$ and $||\nabla D(x,y)||_2^2$ are of small values.

Moreover, this component illuminates the selection criteria for an appropriate proxy model.
While previous studies have highlighted that proxy and target models exhibiting structural and training data similarities tend to facilitate the transferability of adversarial examples across these models, our theoretical insights suggest a more nuanced criterion.
Specifically, a proxy model needs only yield predictions for $x$ that are closely aligned with those of the target model, thereby relaxing the requirements of architectural and data similarities.

\begin{wrapfigure}{r}{6cm}
    \centering
    \includegraphics[width=0.35\textwidth]{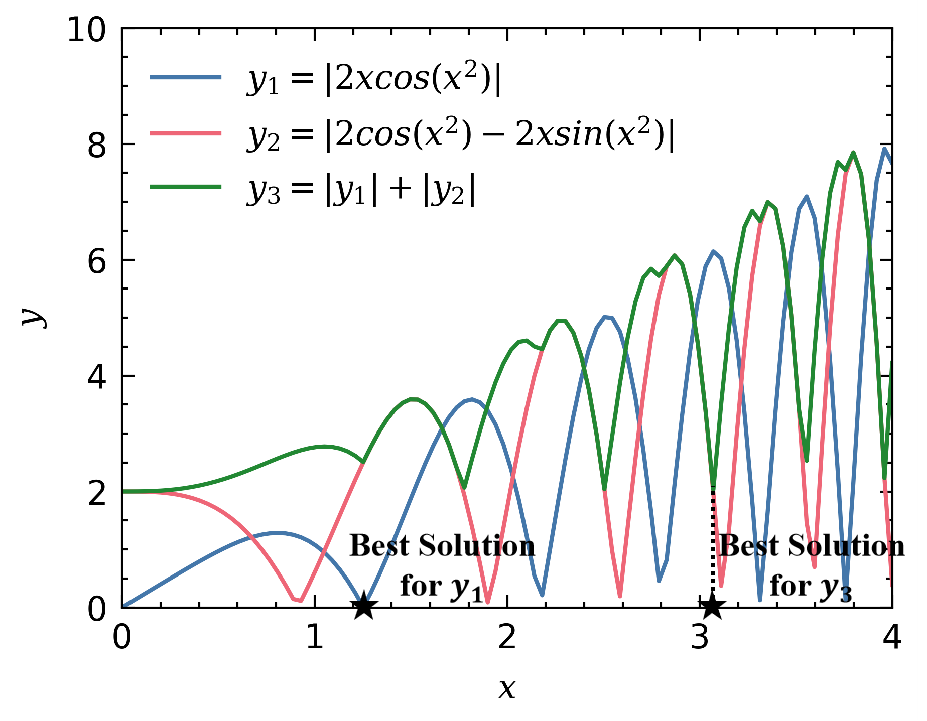}
    \caption{The visualization of the first-order gradient, the second-order gradient of $y=\sin x^2$.
    The black stars symbolize the location where the minimum values of $y_1$ and $y_3$ are achieved.
    }
    \label{demo_case}
\end{wrapfigure}

\textbf{The first-order gradient component and the second-order gradient component.}
Our analysis here challenges the prevailing belief \cite{RAP,flatattack1,flatattack2,flatattack3,flatattack4} that flat adversarial examples necessarily induce improved transferability.
In Theorem \ref{thm1}, the factors that are relevant to the direction of $\delta$ include the first-order and second-order gradient components.
The first-order gradient component serves as a direct measurement of the flatness associated with adversarial examples.
Nonetheless, the presence of the second-order gradient component unveils that transferability is not exclusively contingent on flatness.
In other words, an adversarial example that appears flat but has a higher second-order gradient component might not be optimal for the target model.
To elucidate this point, consider the function $f(x)=\sin(x^2)$ as an example.
Figure \ref{demo_case} presents the norms of first-order gradient $y_1$ and the second-order gradient $y_2$ of $f(x)$, as well as their sum $y_3$.
Upon examination, it is observed that optimizing the norms of $y_1$ alone does not yield optimal solutions for optimizing $y_3$.
This highlights the inadequacy of minimizing only the first-order gradient component in ensuring the minimization of the bound.

\section{The Proposed Approach \sysname}
\label{approach}

\subsection{Overview}

In Section \ref{sec_theory}, the adversarial transferability is decoupled into the local effectiveness term and the transferability-related loss term and we derive a bound for the latter.
It is natural to jointly optimize both the local effectiveness term and the bound for generating more transferable adversarial examples.
The first term maintains the effectiveness of generated adversarial examples against the proxy model while the bound controls the performance degradation upon transfer to the target model, thus achieving better transferability.
However, the bound involves the second-order gradient component that is quite expensive to optimize.
To this end, as shown in Section \ref{sec_our_method_1}, \sysname instead develops a computationally feasible surrogate for the bound that retains optimization integrity, substantiated through theoretical validation.
Section \ref{approach_approximate_solution} introduces an approximate solution to efficiently tackle the optimization target.

\subsection{Optimization Formulation}
\label{sec_our_method_1}


The optimization target of \sysname is formalized as follows:
\begin{equation}
\begin{split}
\delta^* = \underset{\delta}{\arg \min} -\mathcal{L}(F(x+\delta),y) &+ \lambda \mathbb{E}_{\Delta \sim U(-b,b)} \{ || \nabla \mathcal{L}(F(x+\delta+\Delta),y) ||_2 \}, \\
s.t., ||\delta||_{\infty} \leq &\epsilon, b \geq 0,
\end{split}
\label{our_optim}
\end{equation}
where $U(-b,b)$ is uniform distribution between $-b$ and $b$, $\lambda$ is penalty magnitude, and $x+\delta+\Delta$ denotes a sample extracted uniformly from the region around $x+\delta$.
In Equation \ref{our_optim}, we employ $-\mathcal{L}(F(x+\delta),y)$ as the local effectiveness term, while $|| \nabla \mathcal{L}(F(x+\delta+\Delta),y) ||_2$ serves as a surrogate for our bound.
Notice that our proposed surrogate is designed to regulate both the first- and second-order gradient component, excluding the inherent model difference component.
This is because the inherent model difference component remains a constant term given specific perturbation norm constraint and the proxy-target model pair.

We next demonstrate the effectiveness of the surrogate.
The surrogate minimizes the gradients of samples around $x+\delta$, intuitively leading to a minimization of the gradient norms of $x+\delta$ themselves as well.
Our main investigation is determining if the surrogate can effectively minimize the second-order gradient component.
The second-order gradient quantifies the rate at which the first-order gradient changes.
Through penalizing the gradient norms of samples around $x+\delta$, the surrogate naturally encourages the gradient norms of samples around $x+\delta$ towards zero, thereby implicitly moderating the second-order gradient.
Theoretically, it is demonstrated as follows:
\begin{equation}
\label{second_effective_demon}
\begin{split}
&\sum_{i} \ |\nabla^2 \log F(x+\delta)[i,i]| = \sum_{i} \ | \lim_{\mu \rightarrow 0} \frac{\nabla \log F(x+\delta+\mu)[i]-\nabla \log F(x+\delta)[i]}{u}| \\
=  &||\lim_{\mu \rightarrow 0} \{ (\nabla L(F(x+\delta+\mu),y) - \nabla L(F(x+\delta),y)) \cdot \frac{1}{u} \}||_1.
\end{split}
\end{equation}
According to Equation \ref{second_effective_demon}, if the gradient norms of the samples around $x+\delta$ approach zero, the second-order gradient components also tend towards zero.
Thus, the surrogate is capable of effectively regulating both the first- and second-order gradients. 

We see that the optimization target of \sysname bears a high visual similarity to those in some studies \cite{flatattack1,flatattack2} which penalties the gradient norm of $x+\delta$.
Despite this superficial similarity, there exists a fundamental difference in their effects.
As elaborated in Section \ref{sec_theory}, penalizing only the gradient norm of $x+\delta$ is unable to regulate the second-order gradient component.

\subsection{Approximate Solution}
\label{approach_approximate_solution}

The highly non-linear and non-convex nature of $F(\cdot)$ render the analytic solution of Equation \ref{our_optim} to be hardly derived \citep{BIM}.
As a result, standard practice for solving Equation \ref{our_optim} is employing gradient-based optimization methods \citep{auto_pgd}.
However, as shown in Equation \ref{hessian}, the gradients of Equation \ref{our_optim} involve Hessian matrix evaluated at multiple points $x+\delta+\Delta_i$ for $\Delta_i, i=1,\cdots,N$, each of which is troublesome to evaluate in high-dimension spaces \citep{hessian,hessian_appro}.
\begin{equation}
\label{hessian}
\begin{split}
     &- \nabla \mathcal{L}(F(x+\delta),y) + \frac{\lambda}{N} \sum_{i=1}^N \nabla || \nabla \mathcal{L}(F(x+\delta+\Delta_i),y) ||_2 \\
     &= -\nabla \mathcal{L}(F(x+\delta),y)
     + \frac{\lambda}{N} \sum_{i=1}^N H_{x+\delta+\Delta_i}  \frac{\nabla \mathcal{L}(F(x+\delta+\Delta_i),y)}{|| \nabla \mathcal{L}(F(x+\delta+\Delta_i),y) ||_2}.
\end{split}
\end{equation}

To get rid of directly evaluating Hessian matrix in Equation \ref{hessian}, we propose an approximate estimation for $H_{x+\delta+\Delta_i}  \frac{\nabla \mathcal{L}(F(x+\delta+\Delta_i),y)}{|| \nabla \mathcal{L}(F(x+\delta+\Delta_i),y) ||_2}$.
To achieve this, we utilize Taylor expansion on
$\mathcal{L}(F(x+\delta+\Delta_i +\phi),y)$, assuming $\phi$ is small enough for making expansion feasible:
\begin{equation}
\label{taylor_expansion_derive}
\begin{split}
    &\mathcal{L}(F(x+\delta+\Delta_i +\phi),y)= \mathcal{L}(F(x+\delta+\Delta_i),y) + \nabla \mathcal{L}(F(x+\delta+\Delta_i),y) \phi,
\end{split}
\end{equation}
Furthermore, by differentiating both sides of Equation \ref{taylor_expansion_derive}, we obtain:
\begin{equation}
\label{linear_appro}
\begin{split}
   &\nabla \mathcal{L}(F(x+\delta+\Delta_i +\phi),y) 
   = \nabla \mathcal{L}(F(x+\delta+\Delta_i),y) + H_{x+\delta+\Delta_i} \phi, 
\end{split}
\end{equation}
Notice that, the desired item $H_{x+\delta+\Delta_i}  \frac{\nabla \mathcal{L}(F(x+\delta+\Delta_i),y)}{|| \nabla \mathcal{L}(F(x+\delta+\Delta_i),y) ||_2}$ arises if setting $\phi$ along the direction $\frac{\nabla \mathcal{L}(F(x+\delta+\Delta_i),y)}{|| \nabla \mathcal{L}(F(x+\delta+\Delta_i),y) ||_2}$.
By implementing the idea, there is:
\begin{equation}
\label{appro_res}
\begin{split}
     &H_{x+\delta+\Delta_i} \frac{\nabla \mathcal{L}(F(x+\delta+\Delta_i),y)}{|| \nabla \mathcal{L}(F(x+\delta+\Delta_i),y) ||_2} = \\
     &\frac{\nabla \mathcal{L}(F(x+\delta+\Delta_i + k \frac{\nabla \mathcal{L}(F(x+\delta+\Delta_i),y)}{|| \nabla \mathcal{L}(F(x+\delta+\Delta_i),y) ||_2}),y)}{k} - \frac{\nabla \mathcal{L}(F(x+\delta+\Delta_i),y)}{k}.
\end{split}
\end{equation}
Wherein, $\phi=k \frac{\nabla \mathcal{L}(F(x+\delta+\Delta_i),y)}{|| \nabla \mathcal{L}(F(x+\delta+\Delta_i),y) ||_2}$ and $k$ is a small constant to cater $\phi$ being small.
By substituting Equation \ref{appro_res} into Equation \ref{hessian}, we obtain the approximately estimated gradients of Equation \ref{our_optim}.
Compared to directly evaluating Hessian matrix that requires quadratic storage and cubic computation time \citep{hessian,hessian_appro}, our approximate solution only involves first-order gradients (linear computation time) so as to make \sysname more efficient computationally.
Moreover, linear expansion used in Equation \ref{linear_appro} results in an approximation error of $O(k)$~\citep{real_analysis}.

\section{Simulation Experiment}
\label{experiments}

\subsection{Setup}

\textbf{Dataset.}
We randomly select 10000 images from ImageNet.

\begin{table*}[!t]
\caption{The attack success rates (\%) of different attacks on normal models. Best results are in bold.}
\label{main_comp}
\centering
\resizebox{0.99\linewidth}{!}{
\begin{tabular}{@{}c|c|cccccccccccccc@{}}
\toprule
Proxy Model & Method & ResNet50 & DenseNet121 & EfficientNet & InceptionV3 & MobileNetV2 & SqueezeNet & ShuffleNetV2 & ConvNet & RegNet & MNASNet & WideResNet50 & VGG19 & ViT & Swin \\ \midrule
\multirow{8}{*}{ResNet50} 
 & MI & \textbf{100.00} & 87.91 & 76.43 & 68.57 & 66.51 & 86.66 & 78.11 & 60.84 & 72.50 & 81.33 & 90.77 & 81.99 & 43.06 & 55.89 \\
 & NI & \textbf{100.00} & 88.14 & 76.79 & 68.57 & 66.10 & 86.43 & 78.06 & 60.64 & 72.55 & 82.35 & 91.17 & 82.98 & 41.68 & 54.87 \\
 & DI & 99.97 & 80.77 & 68.75 & 69.74 & 66.05 & 77.58 & 72.83 & 49.39 & 65.82 & 79.54 & 82.91 & 76.51 & 40.33 & 46.71 \\
 & TI & \textbf{100.00} & 73.47 & 56.28 & 55.99 & 55.79 & 68.85 & 62.04 & 42.83 & 59.08 & 66.33 & 78.60 & 68.78 & 34.85 & 40.51 \\
 & VT & \textbf{100.00} & 89.62 & 81.63 & 73.47 & 76.86 & 90.21 & 83.83 & 76.28 & 82.83 & 92.86 & 95.51 & 86.12 & 54.77 & 57.87 \\
 & SSA & \textbf{100.00} & 95.51 & 91.56 & 79.62 & 77.40 & 93.60 & 87.60 & 79.90 & 85.99 & 93.60 & 96.53 & 95.13 & 54.21 & 59.64 \\
  & RAP & \textbf{100.00} &  95.01  & 94.78  & 93.81  & 93.88  & 93.87  & 94.64  & 90.80 &  94.54 &  95.25  & 93.97 &  94.92 &  62.62  & 60.77  \\
  & BSR & \textbf{100.00} &  97.02  & 95.16  & 93.58  & 94.19 & 94.52 & 94.88  & 88.46 & 94.64  &  96.70  & 93.62 &  93.80 &  82.32  & 77.05  \\
 & Ours & 99.85 & \textbf{99.69} & \textbf{99.52} & \textbf{98.70} & \textbf{99.52} & \textbf{99.72} & \textbf{99.82} & \textbf{94.34} & \textbf{97.37} & \textbf{99.82} & \textbf{99.44} & \textbf{98.83} & \textbf{93.52} & \textbf{90.98} \\ \midrule
\multirow{8}{*}{ResNet152} 
 & MI & 83.44 & 85.92 & 77.70 & 70.97 & 66.33 & 85.54 & 77.65 & 62.17 & 74.59 & 81.30 & 90.36 & 79.52 & 45.43 & 56.22 \\
 & NI  & 84.26 & 87.32 & 77.70 & 70.38 & 66.38 & 86.10 & 78.06 & 62.88 & 74.67 & 81.48 & 90.89 & 78.70 & 43.93 & 54.31 \\
 & DI & 86.76 & 79.06 & 70.56 & 71.38 & 66.89 & 75.87 & 73.06 & 52.70 & 68.75 & 78.67 & 83.01 & 73.80 & 43.62 & 47.45 \\
 & TI & 84.21 & 73.42 & 58.57 & 57.60 & 56.51 & 68.14 & 62.42 & 45.08 & 61.35 & 66.48 & 78.57 & 66.48 & 37.91 & 41.30 \\
 & VT & 89.54 & 89.57 & 84.67 & 83.62 & 77.98 & 92.07 & 87.76 & 72.42 & 89.29 & 92.5 & 91.38 & 90.87 & 56.38 & 55.87 \\
 & SSA & 94.62 & 95.23 & 93.06 & 88.55 & 78.55 & 96.05 & 93.37 & 76.79 & 95.33 & 93.24 & 94.74 & 93.11 & 57.63 & 59.46 \\
 & RAP & 98.86 &  94.44 &  93.74 &  93.48 &  93.61 &  92.37 &  93.07 &  90.01 &  92.89 &  94.82 &  93.22 &  94.22 &  61.48 &  58.98 \\
 & BSR & 98.05 &  94.62 &  93.55 &  94.08 &  93.75 &  92.42 &  93.30 &  89.75 &  92.52 &  94.04 &  93.64 &  94.82 &  82.22 &  77.86 \\
 & Ours & \textbf{99.85} & \textbf{99.67} & \textbf{99.39} & \textbf{98.95} & \textbf{99.18} & \textbf{99.01} & \textbf{99.64} & \textbf{94.11} & \textbf{97.96} & \textbf{99.77} & \textbf{99.57} & \textbf{98.57} & \textbf{90.19} & \textbf{86.27} \\ \midrule
\multirow{8}{*}{DenseNet121} 
 & MI & 89.01 & 99.97 & 81.35 & 73.95 & 70.33 & 87.88 & 80.18 & 65.00 & 73.65 & 84.41 & 86.63 & 85.64 & 47.17 & 61.10 \\
 & NI & 90.28 & 99.98 & 82.55 & 74.13 & 70.10 & 87.65 & 81.15 & 65.51 & 74.26 & 86.10 & 87.98 & 86.28 & 47.60 & 61.45 \\
 & DI & 82.45 & 99.90 & 73.24 & 72.88 & 70.43 & 78.90 & 76.58 & 51.33 & 66.22 & 81.22 & 78.47 & 79.52 & 43.52 & 49.92 \\
 & TI & 76.33 & 99.95 & 61.38 & 60.15 & 59.01 & 70.41 & 66.33 & 46.15 & 58.78 & 70.84 & 72.76 & 73.60 & 38.88 & 44.21 \\
 & VT & 92.12 & 99.95 & 90.33 & 86.58 & 81.48 & 92.02 & 90.18 & 75.19 & 85.78 & 78.75 & 81.20 & 81.56 & 49.18 & 52.93 \\
 & SSA & 94.82 & 99.97 & 91.86 & 83.67 & 83.67 & 96.66 & 92.76 & 82.18 & 88.04 & 88.11 & 90.48 & 90.13 & 50.92 & 59.64 \\
 & RAP & 97.40 &  99.96 &  95.48 &  95.06 &  94.67 &  97.91 &  94.94 &  90.77 &  92.51 & 96.10 &  95.04 &  94.43 &  63.13 &  61.65 \\
 & BSR & 98.24 &  99.99  & 94.68  & 94.24  & 94.56 & 96.27  & 94.96  & 90.52 & 93.01  & 95.63  & 89.20 &  95.42 &  81.52  & 79.25  \\
 & Ours & \textbf{99.69} & 99.87 & \textbf{99.34} & \textbf{98.32} & \textbf{99.52} & \textbf{99.57} & \textbf{99.82} & \textbf{94.41} & \textbf{97.37} & \textbf{99.77} & \textbf{99.08} & \textbf{99.03} & \textbf{93.70} & \textbf{91.98} \\ \midrule
\multirow{8}{*}{DenseNet201}
 & MI & 90.41 & 95.89 & 83.16 & 74.90 & 70.56 & 87.53 & 80.69 & 69.21 & 76.33 & 85.08 & 87.60 & 83.34 & 51.28 & 64.31 \\
 & NI & 91.20 & 96.56 & 83.19 & 75.48 & 70.20 & 87.32 & 80.87 & 70.18 & 77.24 & 86.25 & 88.55 & 84.67 & 50.48 & 64.54 \\
 & DI & 84.57 & 90.28 & 76.02 & 75.71 & 70.74 & 77.68 & 77.76 & 56.71 & 70.99 & 82.53 & 81.38 & 78.57 & 47.78 & 53.67 \\
 & TI & 78.39 & 88.11 & 63.85 & 62.37 & 59.67 & 70.00 & 66.61 & 49.21 & 63.29 & 71.30 & 74.11 & 70.74 & 41.53 & 46.38 \\
 & VT & 93.44 & 95.91 & 83.06 & 87.40 & 80.71 & 89.92 & 93.04 & 79.92 & 84.89 & 79.16 & 82.12 & 79.16 & 42.37 & 55.66 \\
 & SSA & 96.20 & 97.27 & 92.42 & 84.11 & 82.88 & 94.59 & 90.31 & 77.93 & 90.33 & 87.42 & 89.82 & 87.70 & 42.50 & 62.12 \\
 & RAP & 96.01 &  98.99 &  94.42 &  95.15 &  93.27 &  97.56 &  93.58 &  89.05 &  91.04 &  95.64 &  93.74 &  92.52 &  62.66 &  61.31 \\
 & BSR & 96.85 &  97.94 &  94.24 &  95.08 &  93.26 &  96.26 &  93.54 &  92.24 &  92.30 &  95.16 &  94.56 &  92.46 &  81.02 &  79.51 \\
 & Ours & \textbf{99.72} & \textbf{99.90} & \textbf{99.67} & \textbf{98.83} & \textbf{99.23} & \textbf{99.31} & \textbf{99.85} & \textbf{96.96} & \textbf{98.34} & \textbf{99.77} & \textbf{99.62} & \textbf{99.16} & \textbf{93.54} & \textbf{92.85} \\ \bottomrule
\end{tabular}
}
\end{table*}

\textbf{Models.}
We consider 14 models including ResNet50, DenseNet121, MobileNetV2, EfficientNet, VGG19, InceptionV3, WideResNet50, MNASNet, RegNet, ShuffleNetV2, SqueezeNet, ConvNet, ViT, and Swin.
with the first twelve being convolutional networks and the final two based on transformer architectures.
Moreover, we also validate the performance of \sysname on secured models including adversarial training with $L_2$ \citep{l2_training_models} and $L_{\infty}$ \citep{adv_training_models} constraint as well as robust training with Styled ImageNet (SIN) and the mixture of Styled and natural ImageNet (SIN-IN) \citep{SIN}.

\textbf{Baselines.}
Nine state-of-the-art attacks are used as competitors for \sysname: BIM \citep{BIM}, DI \citep{DI}, MI \citep{MI}, NI \citep{NI_SI}, TI \citep{TI}, VT \citep{VR}, SSA \citep{SSA}, RAP \citep{RAP}, BSR \citep{BSR}, and Self-Universality \citep{self_univer}.
Self-Universality is a targeted attack and thus we only report its performance in targeted setting.

\textbf{Evaluation metric.}
Attack success rate (ASR, $\uparrow$) is used as the evaluation metric that is defined as the misclassified rate of adversarial examples by target models.

\textbf{Hyperparameter configurations.}
For baselines, we set hyperparameters used in their original papers.
For \sysname, we set $\lambda=5, b=16, k=0.05, N=10$.
Moreover, for all methods, we set iteration of 20\footnote{Since our approximation involves Hessian matrix that is known to accelerate convergence, we increase the number of iterations to ensure the convergence of all attacks. See Appendix \ref{appendix_varying_iter} for attack results with iteration of 10. Overall, the increased iterations did not significantly affect the attack performance (less than 1\%).}, $\epsilon$ of 16, and step size of 1.6.

\begin{table*}[!t]
\caption{The attack success rates (\%) of different methods on secured models. Three different robust training methods are considered: adversarial training with $L_2$ perturbation ($L2-\{0.03 \sim 5 \}$) \citep{l2_training_models} and $L_{\infty}$ perturbation (AdvIncV3 and EnsAdvIncResV2) \citep{adv_training_models}, robust training with Styled ImageNet (SIN) and the mixture of Styled and natural ImageNet (SIN-IN) \citep{SIN}. The best results are in bold.}
\label{adv_trained_comp}
\centering
\resizebox{0.85\linewidth}{!}{
\begin{tabular}{@{}c|c|ccccccccccc@{}}
\toprule
\multicolumn{1}{l|}{Proxy Model} & Method & AdvIncV3 & EnsAdvIncResV2 & SIN & SIN-IN & L2-0.03 & L2-0.05 & L2-0.1 & L2-0.5 & L2-1 & L2-3 & L2-5 \\ \midrule
\multirow{10}{*}{ResNet50} & BIM & 41.81 & 33.52 & 50.51 & 51.56 & 63.67 & 64.06 & 62.81 & 61.10 & 63.98 & 68.27 & 74.13 \\
 & MI & 58.78 & 43.24 & 75.46 & 96.05 & 82.83 & 83.47 & 80.31 & 74.74 & 75.00 & 75.08 & 77.98 \\
 & NI & 58.11 & 43.11 & 75.13 & 96.79 & 83.57 & 83.70 & 79.97 & 74.52 & 75.15 & 75.33 & 78.11 \\
 & DI & 54.34 & 45.03 & 75.84 & 92.04 & 78.27 & 78.19 & 75.61 & 69.57 & 70.94 & 71.53 & 75.74 \\
 & TI & 48.52 & 39.72 & 66.35 & 88.27 & 75.51 & 75.05 & 72.76 & 67.04 & 69.21 & 70.87 & 75.61 \\
 & VT & 59.72 & 50.89 & 69.03 & 94.06 & 77.27 & 77.22 & 74.21 & 68.24 & 69.80 & 70.92 & 75.48 \\
 & SSA & 60.13 & 50.48 & 70.31 & 97.63 & 80.61 & 80.46 & 76.84 & 69.06 & 70.28 & 71.25 & 75.69 \\
 & RAP  & 60.17 & 51.02 & 84.34 & 97.94 & 81.72 & 81.43 & 80.22 & 75.49 & 75.06 & 72.48 & 76.05 \\
 & BSR  & 64.02 & 56.35 & 88.47 & 98.21 & 85.34 & 85.57 & 83.09 & 78.31 & 78.31 & 76.28 & 80.61 \\ 
 & Ours & \textbf{74.57} & \textbf{72.81} & \textbf{99.52} & \textbf{99.90} & \textbf{96.20} & \textbf{96.45} & \textbf{96.48} & \textbf{93.65} & \textbf{91.76} & \textbf{88.19} & \textbf{87.24} \\ \midrule
\multirow{10}{*}{ResNet152} & BIM & 41.84 & 33.85 & 48.95 & 42.14 & 63.11 & 63.47 & 62.76 & 60.92 & 64.11 & 68.19 & 74.06 \\
 & MI & 58.78 & 45.15 & 71.40 & 82.58 & 82.55 & 82.96 & 80.13 & 74.87 & 75.23 & 75.36 & 78.21 \\
 & NI & 58.98 & 44.74 & 71.79 & 83.83 & 82.70 & 83.04 & 80.08 & 74.80 & 75.13 & 75.28 & 78.14 \\
 & DI & 54.97 & 47.58 & 72.50 & 80.00 & 77.27 & 76.86 & 75.38 & 69.46 & 70.26 & 71.56 & 75.59 \\
 & TI & 49.34 & 41.48 & 63.11 & 72.76 & 74.21 & 74.44 & 72.22 & 66.61 & 69.06 & 71.35 & 75.48 \\
 & VT & 51.22 & 43.42 & 65.10 & 78.62 & 76.58 & 76.07 & 73.44 & 68.11 & 69.72 & 71.02 & 75.41 \\
 & SSA & 51.30 & 41.94 & 67.68 & 84.62 & 79.34 & 79.31 & 76.61 & 69.29 & 70.20 & 71.20 & 75.54 \\
 & RAP  & 52.38 & 43.36 & 76.19 & 85.47 & 81.12 & 81.21 & 79.54 & 75.42 & 74.95 & 72.16 & 75.39 \\
 & BSR  & 53.28 & 45.56 & 80.95 & 88.37 & 84.95 & 84.84 & 82.78 & 77.57 & 77.60 & 75.61 & 80.42 \\
 & Ours & \textbf{71.66} & \textbf{79.46} & \textbf{99.21} & \textbf{99.80} & \textbf{95.64} & \textbf{95.43} & \textbf{95.15} & \textbf{91.76} & \textbf{90.05} & \textbf{85.36} & \textbf{84.72} \\ \midrule
\multirow{10}{*}{DenseNet121} & BIM & 41.99 & 34.08 & 48.80 & 40.26 & 63.09 & 63.11 & 62.55 & 60.92 & 64.01 & 68.27 & 74.16 \\
 & MI & 59.31 & 45.03 & 71.25 & 79.44 & 81.71 & 83.27 & 80.08 & 74.64 & 76.02 & 75.77 & 78.65 \\
 & NI & 59.03 & 45.08 & 72.07 & 80.31 & 81.53 & 83.14 & 79.80 & 74.87 & 75.69 & 75.79 & 78.57 \\
 & DI & 55.00 & 47.19 & 73.21 & 77.19 & 76.99 & 77.07 & 75.94 & 69.31 & 70.69 & 71.68 & 75.69 \\
 & TI & 49.13 & 40.71 & 62.24 & 67.58 & 73.62 & 73.57 & 71.96 & 67.60 & 68.85 & 71.10 & 75.38 \\
 & VT & 60.26 & 51.73 & 64.26 & 75.28 & 76.28 & 76.38 & 74.21 & 68.55 & 69.97 & 71.05 & 75.54 \\
 & SSA & 61.84 & 51.86 & 69.08 & 84.06 & 79.64 & 79.85 & 76.94 & 69.67 & 70.51 & 71.45 & 75.77 \\
 & RAP  & 62.17 & 53.02 & 75.62 & 84.92 & 81.62 & 81.10 & 79.65 & 75.45 & 74.78 & 72.00 & 75.64 \\
 & BSR  & 64.02 & 56.35 & 81.14 & 88.25 & 85.14 & 84.69 & 82.24 & 78.22 & 77.39 & 76.18 & 79.78 \\
 & Ours & \textbf{72.19} & \textbf{79.64} & \textbf{99.16} & \textbf{99.64} & \textbf{95.31} & \textbf{95.69} & \textbf{95.56} & \textbf{92.65} & \textbf{90.82} & \textbf{86.63} & \textbf{85.71} \\ \midrule
\multirow{10}{*}{DenseNet201} & BIM & 42.07 & 34.18 & 48.93 & 40.99 & 62.88 & 63.62 & 62.78 & 60.94 & 64.11 & 68.42 & 74.13 \\
 & MI & 59.92 & 45.20 & 71.33 & 81.53 & 82.98 & 83.19 & 80.03 & 75.54 & 76.20 & 75.71 & 78.60 \\
 & NI & 59.90 & 44.41 & 72.27 & 82.42 & 82.55 & 83.06 & 80.18 & 75.28 & 75.79 & 75.66 & 78.21 \\
 & DI & 56.53 & 48.70 & 74.69 & 80.03 & 77.12 & 77.45 & 76.73 & 70.05 & 71.43 & 71.71 & 75.82 \\
 & TI & 50.08 & 41.20 & 62.68 & 69.34 & 74.21 & 74.29 & 72.60 & 67.68 & 69.67 & 71.02 & 75.64 \\
 & VT & 61.76 & 52.86 & 65.23 & 76.51 & 76.66 & 76.86 & 74.59 & 68.67 & 70.26 & 71.07 & 75.64 \\
 & SSA & 62.30 & 53.76 & 69.59 & 85.05 & 79.64 & 79.39 & 76.76 & 70.20 & 70.92 & 71.38 & 75.99 \\
 & RAP  & 63.17 & 54.02 & 75.70 & 85.27 & 80.76 & 81.06 & 80.16 & 74.74 & 74.16 & 71.70 & 75.33 \\
 & BSR  & 65.02 & 56.35 & 81.41 & 88.03 & 85.20 & 85.10 & 83.06 & 77.45 & 77.56 & 76.17 & 80.10 \\
 & Ours & \textbf{70.99} & \textbf{78.21} & \textbf{98.98} & \textbf{99.67} & \textbf{95.41} & \textbf{95.94} & \textbf{94.92} & \textbf{91.56} & \textbf{89.44} & \textbf{85.51} & \textbf{84.57} \\ \bottomrule
\end{tabular}
}
\end{table*}

\begin{table*}[t]
\caption{The targeted attack success rates of different methods. The proxy model is ResNet50.}
\label{targeted_res}
\centering
\resizebox{0.85\linewidth}{!}{
\begin{tabular}{@{}c|cccccccc@{}}
\toprule
Method & DenseNet121 & EfficientNet & InceptionV3 & ConvNet & WideResNet50 & VGG19 & ViT & Swin \\ \midrule
MI & 4.20 & 0.60 & 0.00 & 0.20 & 5.20 & 0.80 & 0.00 & 0.20 \\
DI & 1.60 & 0.20 & 0.20 & 0.00 & 2.00 & 0.70 & 0.00 & 0.10 \\
VT & 4.70 & 1.10 & 0.50 & 1.10 & 5.40 & 1.30 & 0.40 & 0.40 \\
SSA & 5.40 & 2.30 & 1.50 & 0.70 & 6.60 & 3.60 & 0.80 & 1.20 \\
RAP & 18.60 & 13.90 & 8.20 & 6.70 & 12.20 & 7.10 & 7.30 & 5.80 \\
BSR & 19.40 & 14.00 & 8.60 & 7.90 & 14.50 & 9.30 & 7.70 & 6.50 \\
Self-Universality & 22.00 & 18.00 & 9.60 & 8.70 & 15.70 & 8.50 & 9.70 & 6.70 \\
Ours & \textbf{31.90} & \textbf{25.20} & \textbf{14.30} & \textbf{12.20} & \textbf{22.60} & \textbf{13.20} & \textbf{13.40} & \textbf{9.80} \\ \bottomrule
\end{tabular}
}
\end{table*}

\begin{figure*}[!t]
    \centering
    \includegraphics[width=0.32\linewidth]{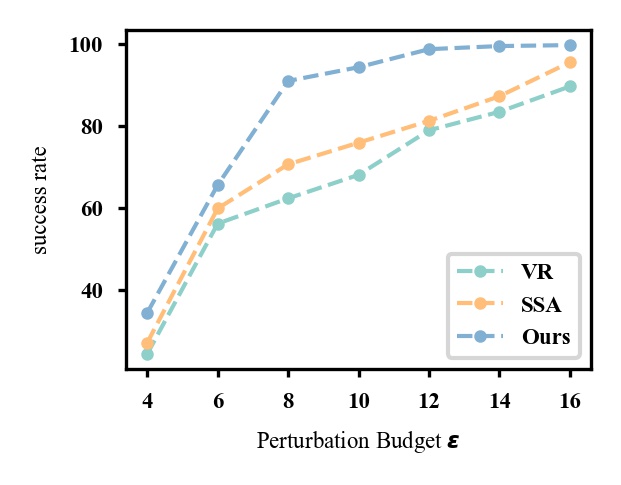}
    \includegraphics[width=0.32\linewidth]{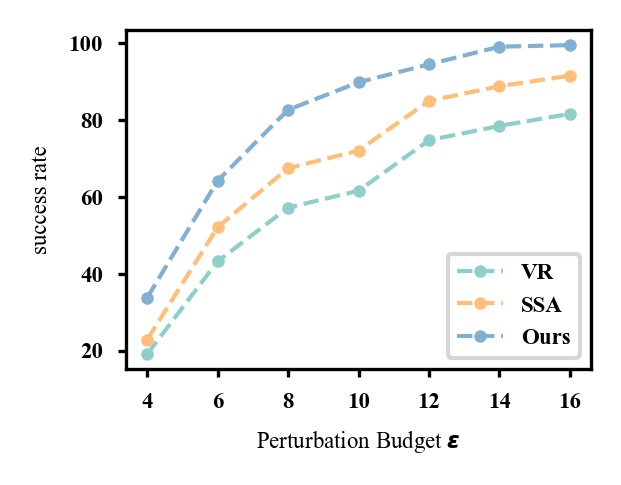}
    \includegraphics[width=0.32\linewidth]{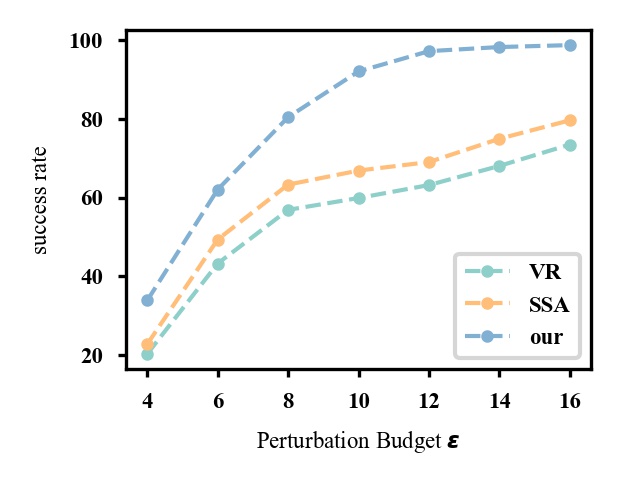}
    \caption{
    The attack effectiveness of \sysname with varying perturbation budgets. 
    The adversarial examples are crafted on ResNet50.
    }
    \label{impact_epsilon}
\end{figure*}

\subsection{Attack Results}

\textbf{Attack results on normal models.}
Here we examine the attack effectiveness of our method on both convolutional neural networks and transformer-like neural networks and Table \ref{main_comp} reports the attack results.
The following observation can be made.
Simply speaking, as shown in Table \ref{main_comp}, our method defeats all the baselines by a significant margin and is the winning attack.
For instance, when employing ResNet50 as the proxy model, our attack witnesses an average gain of roughly 9\% on attack success rate, even when compared to the state-of-the-art transfer-based attack SSA.

\textbf{Attack results on secured models.}
We evaluate the effectiveness of \sysname on models defended by robust training.
We stick to employing undefended models as proxy models and this is a more challenging setting, given that proxy and target models are more divergent.
The attack results are reported in Table \ref{adv_trained_comp} and we draw the following conclusions.
Similar to the attack results on normal models, \sysname still can craft more transferable adversarial examples against robust models.
In particular, as can be seen in Table \ref{adv_trained_comp}, \sysname enhances ASRs over SSA by a large margin.

\textbf{Targeted attack results.}
We also investigate the attack effectiveness of \sysname for targeted attacks, which is a more challenging attack setting compared to untargeted attack setting.
Following the evaluation setting for targeted attacks \citep{targeted_attack}, Table \ref{targeted_res} reports the targeted attack results of different attack methods.
As can be seen in Table \ref{targeted_res}, the attack effectiveness of \sysname on the targeted setting is so striking, even when compared to Self-Universality, which is specifically designed for targeted attacks.
compared with baselines.

\textbf{Attack results with varying perturbation budgets.}
Small perturbation budgets $\epsilon$ raise the difficulty in conducting transfer-based attacks and we further evaluate the attack effectiveness of different methods with varying perturbation budgets.
Figure \ref{impact_epsilon} illustrates the ASRs of VT, SSA, and \sysname.
In short, \sysname still dominates all settings in terms of ASRs.

\section{Evaluation in Real World Applications}
\label{real_eval}

This section evaluates the performance of \sysname against real applications, which is more challenging but also leads to a more reliable evaluation due to the following three reasons:
\begin{itemize}
    \item \textbf{Application architecture and complexity.} Real-world applications are likely to employ sophisticated models that are difficult to replicate in a controlled research environment. Besides, such applications probably incorporate practical defenses.
    \item \textbf{Training setting.} Publicly available models mostly share similar training recipes (ImageNet). However, the recipes of real industry environments are far more complicated.
    \item \textbf{Structure of the output.} Real-world systems often output multiple hierarchical labels with associated confidences, instead of logits. Along this, the output space of real applications is vastly larger than that of proxy models, i.e., inconsistency in label space.
\end{itemize}

\begin{table*}[!t]
\centering
\caption{The scoring for the effectiveness of adversarial examples against real-world applications.
We randomly extract 100 samples from ImageNet and generate adversarial examples for them using \sysname and ResNet50.
We enlist a volunteer to assess the consistence between the image contents with the predictions made by applications.
A lower rating reflects a higher effectiveness of the attack.
}
\label{real_world_res}
\resizebox{0.95\linewidth}{!}{
\begin{tabular}{@{}ccccccccc@{}}
\toprule
Score ($\downarrow$) & Classification & Object Detection & Google Search & Bing Search & Yandex Search & Baidu Search & GPT-4 & Claude3 \\ \midrule
5 & 1 & 3 & 0 & 0 & 0 & 0 & 2 & 0 \\ 
4 & 7 & 21 & 10 & 6 & 5 & 4 & 15 & 12 \\
3 & 13 & 7 & 18 & 11 & 13 & 4 & 27 & 26 \\
2 & 9 & 4 & 16 & 21 & 17 & 10 & 30 & 30 \\
1 & 70 & 65 & 56 & 62 & 65 & 82 & 26 & 32 \\
\bottomrule
\end{tabular}
}
\end{table*}

\textbf{Google MLaaS platforms.}
We attack \href{https://cloud.google.com/vision/docs/drag-and-drop}{Google Cloud Vision Application} including image classification and object detection, which is recognized as one of the most advanced AI services.
We craft 100 adversarial examples against the application.
We first collect the application's responses to these examples and ask a volunteer to score the consistency between these examples with the corresponding responses \footnote{See Appendix \ref{ethic_statement} for more information on the recruitment of volunteers and the evaluation procedure.}.
The scoring ranges from 1 (totally wrong) to 5 (precise), with higher scores indicating weaker attack performance.
See the appendix \ref{appendix_score_implication} for further details about the scoring process.
Table \ref{real_world_res} shows the striking effectiveness of \sysname against Google Service, while Appendix \ref{appendix_visual_real_world} provides visualization results.
The ASRs for image classification and object detection are around 70\% and 80\%, respectively, considering scores of 1 or 2 as a successful attack.
Besides, we find that ineffective adversarial examples often involve entities of persons, a label not covered in ImageNet.
As a result, \sysname is not guided to corrupt the features of "Person", leading to the ineffectiveness of these adversarial examples.

\textbf{Reverse image search engines.}
Given an image of interest, reverse image search engines enable searching for the most similar ones and creates great convenience and benefits.
The task is remarkably different from image classification and object detection and here we test the effectiveness of \sysname against search engines.
We attack Top-4 search engines suggested by \href{https://www.reverseimagesearch.com/}{the site}, including \href{https://images.google.com/}{Google}, \href{https://cn.bing.com/visualsearch}{Bing}, \href{https://yandex.com/images/}{Yandex}, and \href{https://image.baidu.com/}{Baidu}.
We reuse adversarial ones crafted for Google Service and score their effectiveness ranging from 1 to 5, inversely related to the similarity of retrieved images to the original image.
Table \ref{real_world_res} reports attack results and Appendix \ref{appendix_visual_real_world} shows the images retrieved by four search engines for original and adversarial ones.
Four engines present notable vulnerabilities against adversarial examples by \sysname, particularly Baidu Picture Search, which returns completely unrelated images.

\begin{figure}[!t]
  \centering
  \begin{minipage}{.3\textwidth}
    \centering
    \includegraphics[width=0.4\textwidth]{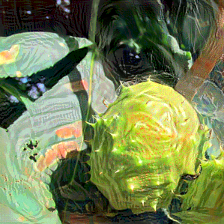}
    \caption{Example.}
    \label{fig:test1}
  \end{minipage}%
  \begin{minipage}{.32\textwidth}
    \centering
    \raggedright
    \small
    \textbf{GPT-4 Response:} The image appears to be a digital painting or an animated, glitch-like effect applied to a photograph of a green cabbage or lettuce being chopped with a knife. The colors are vivid, and the motion suggests that the knife is in the act of slicing through the vegetable.
  \end{minipage}
  \hspace{+0.2cm}
  \begin{minipage}{.32\textwidth}
    \centering
    \vspace{-0.04cm}
    \raggedright
    \small
    \textbf{Claude3 Response:} This image shows various plastic bags filled with different types of produce and vegetables. The bags contain items like lettuce or leafy greens, carrots, and what appears to be herbs or greens of some kind. The plastic bags allow the contents to be visible from the outside.
  \end{minipage}
\end{figure}

\textbf{Multi-modal AI Platforms.}
Multimodal AI platforms, which process and integrate various types of data, are inherently more difficult to attack due to their complexity and their ability for contextual integration across multiple modalities.
We evaluate the effectiveness of our method on the multimodal AI platforms of OpenAI and Amazon, specifically on GPT-4 and Claude3, respectively.
Table \ref{real_world_res} reports the numerical results where \sysname achieves ASRs as high as 56\% and 62\% for GPT-4 and Claude3 if scores of 1 or 2 are deemed as a successful attack.
Specifically, Figure 1 provides an example wherein both GPT-4 and Claude3 fail to accurately classify the green fruit and mistakenly perceive a knife and plastic bag instead.

\section{Ending Remark}
\label{conclusion}
In this paper, we established a bound for adversarial transferability and demonstrated that flat adversarial examples do not inherently enjoy improved transferability.
Drawing upon our theoretical insights, we developed \sysname, which optimizes both the local effectiveness term and a surrogate of the transfer-related loss term to generate adversarial examples.
We conducted extensive experiments to validate the effectiveness of \sysname across standard benchmarks and multiple real-world applications.
It is our hope that our theoretical analysis will inspire further research that delves deeper into adversarial transferability, driving forward the development of more sophisticated and effective adversarial attacks and defenses.

\section{Acknowledgments}
This work was supported by the National Natural Science Foundation of China under grant number 62202170 and the Alibaba Group through the Alibaba Innovation Research Program.

\bibliographystyle{plainnat}
\bibliography{reference}

\clearpage

\appendix
\section{Proof of Theorem \ref{thm1}}
\label{proof_theory}

We first demonstrate a bound on our transfer loss term $D(x+\delta,y)=\mathcal{L}(F'(x+\delta),y) - \mathcal{L}(F(x+\delta),y)$.
By employing Taylor expansion on $D(x+\delta,y) =\mathcal{L}(F'(x+\delta),y) - \mathcal{L}(F(x+\delta),y)$, we derive the following expression:
\begin{equation}
\begin{split}
&D(x, y) - D(x+\delta, y) = - \nabla L(F'(x+\delta),y)^\top \delta + \nabla L(F(x+\delta),y)^\top \delta.
\end{split}
\end{equation}
Let us denote the underlying distribution of $x$ as $p(x)$.
We employ cross-entropy loss as the loss function.
We use $F(x)[y]$ and $F'(x)[y]$ to denote the prediction probabilities of classifying input $x$ as $y$.
We consider minimizing $||\nabla L(F'(x+\delta),y)^\top \delta - \nabla L(F(x+\delta),y)^\top||_2^2$ over $p(x)$.
There is:
\begin{equation}
\label{appendix_equ_9}
\begin{split}
  & \int p(x) || ( \nabla \log F'(x+\delta)[y] - \nabla \log F(x+\delta)[y]^\top \delta ||_2^2 dx \\
  & \leq \int p(x) ||\delta||_2^2 || \nabla \log F'(x+\delta)[y] - \nabla \log F(x+\delta)[y] ||_2^2 dx \quad (by \ ||ab|| \leq ||a|| \ ||b||) \\
  & = \int p(x) ||\delta||_2^2 ( || \nabla \log F'(x+\delta)[y] ||_2^2 + || \nabla \log F(x+\delta)[y] ||_2^2 ) dx \\
  & - 2 \int p(x) ||\delta||_2^2 \nabla \log F'(x+\delta)[y]^\top \nabla \log F(x+\delta)[y]  dx.
\end{split}
\end{equation}
The following proof is divided into two main steps.
The first step involves deriving a bound on the second integral. The second step focuses on deriving a bound on $\int  p(x) ||\nabla \log F'(x+\delta)[y]||_2^2 dx$.
Finally, we combine the results from both steps to obtain the bound for the equation above.

\textbf{Step I: we handle the second integral of the above expression here: $\int p(x) ||\delta||_2^2 \nabla \log F'(x+\delta)[y]^\top \nabla \log F(x+\delta)[y]  dx$.}
We have:
\begin{equation}
\label{appendix_equ_10}
\begin{split}
  & \int p(x) ||\delta||_2^2 \nabla \log F'(x+\delta)[y]^\top \nabla \log F(x+\delta)[y]  dx \\
  &= \int \frac{p(x)}{F'(x+\delta)[y]} ||\delta||_2^2 \nabla F'(x+\delta)[y]^\top \nabla \log F(x+\delta)[y]  dx \\
  & \geq \int p(x) ||\delta||_2^2 \nabla F'(x+\delta)[y]^\top \nabla \log F(x+\delta)[y] dx \quad (use \  F'(x+\delta)[y] \leq 1)\\
  &= \sum_{i} \int p(x) ||\delta||_2^2 \nabla F'(x+\delta)[y][i] \cdot \nabla \log F(x+\delta)[y][i] dx \\
  &= \sum_{i} p(x)||\delta||_2^2 F'(x+\delta)[y] \cdot \nabla \log F(x+\delta)[y][i]|_{- \infty}^{+ \infty} \\
  &- \sum_{i} \int p(x)||\delta||_2^2 F'(x+\delta)[y] \nabla^2 \log F(x+\delta)[y][i,i] dx \\
  &= - \sum_{i} \int p(x)||\delta||_2^2 F'(x+\delta)[y] \nabla^2 \log F(x+\delta)[y][i,i] dx \quad (use \ p(\infty)=0)  \\
  &\geq - \sum_{i} \int p(x)||\delta||_2^2 F(x+\delta)[y] \nabla^2 \log F(x+\delta)[y][i,i] dx.
\end{split}
\end{equation}
Since adversarial examples generated on the proxy model are often also somewhat effective against the target model, this implies that $\nabla F'(x+\delta)[y]^\top \nabla \log F(x+\delta)[y] \geq 0$.
Furthermore, by using $F'(x+\delta)[y] \leq 1$, we can derive the first inequality in the above equation.
$[i]$ and $[i,i]$ refer to $i$-th element in gradient and element in the $i$-th row and $i$-th column of Hessian matrix, respectively.
From the fourth line to the fifth line, we use integration by parts.
Then, we suggest $p(+\infty)=p(-\infty)=0$, which is natural in the distribution of interest.
In the last line, we use $F'(x+\delta)[y] \leq F(x+\delta)[y]$ and $\nabla^2 \log F(x+\delta)[y][i,i] \leq 0$.
Since $\delta$ is crafted for the proxy model, it naturally follows that $F'(x+\delta)[y] \leq F(x+\delta)[y]$, i.e., Assumption \ref{assumpt4}.
Moreover, the second derivative of $log(\cdot)$ is negative, and the second derivative of $F(x+\delta)[y][i,i]$ is 0, which leads to $\nabla^2 \log F(x+\delta)[y][i,i] \leq 0$.
The second derivative being zero can be derived from Assumption \ref{assumpt5}, which is easily verifiable.

\textbf{Step 2: now we derive the bound of $\int  p(x) ||\nabla \log F'(x+\delta)[y]||_2^2 dx$.}
Notice that here we have:
\begin{equation}
\label{identity_for_target_model}
\begin{split}
        &\int  p(x+\delta) (||\nabla \log F'(x+\delta)[y]||_2^2 - ||\nabla \log F(x+\delta)[y]||_2^2) dx \\
        &= \int  p(x) (||\nabla \log F'(x)[y]||_2^2 - ||\nabla \log F(x)[y]||_2^2) dx.
\end{split}
\end{equation}

Based on Equation \ref{identity_for_target_model}, we obtain:
\begin{equation}
\begin{split}
        &\int  p(x+\delta) ||\nabla \log F'(x+\delta)[y]||_2^2 dx \\
        &= \int  p(x+\delta) ||\nabla \log F(x+\delta)[y]||_2^2 dx \\
        &+ \int  p(x) (||\nabla \log F'(x)[y]||_2^2 - ||\nabla \log F(x)[y]||_2^2) dx.
\end{split}
\end{equation}

For $\int  p(x) (||\nabla \log F'(x)[y]||_2^2 - ||\nabla \log F(x)[y]||_2^2) dx$, by using $||a||-||b|| \leq |\ ||a||-||b||\ | \leq ||a-b||$, we have:
\begin{equation}
\begin{split}
    &\int  p(x) (||\nabla \log F'(x)[y]||_2^2 - ||\nabla \log F(x)[y]||_2^2) dx \\
    &\leq \int  p(x) ||\nabla (\log F'(x)[y] - \log F(x)[y])||_2^2 dx.
\end{split}
\end{equation}

Thus, there is:
\begin{equation}
\label{final_derive}
\begin{split}
        &\int  p(x+\delta) ||\nabla \log F'(x+\delta)[y]||_2^2 dx \\
        &\leq \int  p(x+\delta) ||\nabla \log F(x+\delta)[y]||_2^2 dx \\
        &+ \int  p(x) ||\nabla (\log F'(x)[y] - \log F(x)[y])||_2^2 dx.
\end{split}
\end{equation}

Due to $p(x+\delta) \leq p(x)$, there is:
\begin{equation}
\begin{split}
        &\int  p(x+\delta) ||\nabla \log F'(x+\delta)[y]||_2^2 dx \\
        &\leq \int  p(x) ||\nabla \log F(x+\delta)[y]||_2^2 dx \\
        &+ \int  p(x) ||\nabla (\log F'(x)[y] - \log F(x)[y])||_2^2 dx.
\end{split}
\end{equation}

Furthermore, we let:
\begin{equation}
\begin{split}
        C &= \max \frac{p(x)}{p(x+\delta)} \\
        &= \max \frac{p(x)}{p(x)+\nabla p(x)^\top \delta} \quad (\text{notice} \ \nabla p(x)^\top \delta \geq 0)\\
        &= \max \frac{B_1}{B_1+\nabla p(x)^\top \delta} \\
        &= \max \frac{B_1}{B_1+B_2 ||\delta||_2} \\
        &= \frac{B_1}{B_1+B_2 ||\delta||_2}.
\end{split}
\end{equation}

Notice that:
\begin{equation}
\begin{split}
        p(x+\delta) = p(x+\delta)\frac{p(x)}{p(x)} \geq p(x) \min \frac{p(x+\delta)}{p(x)} = \frac{B_1+B_2 ||\delta||_2}{B_1}p(x).
\end{split}
\end{equation}

For Equation \ref{final_derive}, we have:
\begin{equation}
\begin{split}
        &\int  p(x) ||\nabla \log F'(x+\delta)[y]||_2^2 dx \\
        &\leq C \int  p(x) ||\nabla \log F(x+\delta)[y]||_2^2 dx \\
        &+ C \int  p(x) ||\nabla (\log F'(x)[y] - \log F(x)[y])||_2^2 dx.
\end{split}
\end{equation}

\textbf{Combining results.}
Combining Equation \ref{appendix_equ_9} and \ref{appendix_equ_10}, we have:
\begin{equation}
\label{eq_99}
\begin{split}
  & \int p(x) || ( \nabla \log F'(x+\delta)[y] - \nabla \log F(x+\delta)[y] )^\top \delta ||_2^2 dx \\
  & \leq \int p(x) ||\delta||_2^2 ( || \nabla \log F'(x+\delta)[y] ||_2^2 + || \nabla \log F(x+\delta)[y] ||_2^2 ) dx \\
  &+ 2 \sum_{i} \int p(x)||\delta||_2^2 F(x+\delta)[y] \nabla^2 \log F(x+\delta)[y][i,i] dx \\
  & \leq (1+C) \int p(x) ||\delta||_2^2 || \nabla \log F(x+\delta)[y] ||_2^2  dx \\
  &+ 2 \sum_{i} \int p(x) ||\delta||_2^2 |\nabla^2 \log F(x+\delta)[y][i,i]| dx \\
  &+ C \int  p(x) ||\delta||_2^2 ||\nabla (\log F'(x)[y] - \log F(x)[y])||_2^2 dx.
\end{split}
\end{equation}
In the second line of Equation \ref{appendix_equ_10}, we use $F(x+\delta) \leq 1$.
Then, according to $D(x, y) - D(x+\delta, y) = - \nabla L(F'(x+\delta),y)^\top \delta + \nabla L(F(x+\delta),y)^\top \delta$, we have 
\begin{equation}
    \mathbb{E}_{p(x)} ||D(x+\delta,y) ||_2^2 \leq \mathbb{E}_{p(x)} ||D(x, y)||_2^2 + \mathbb{E}_{p(x)} ||\nabla L(F'(x+\delta),y)^\top \delta + \nabla L(F(x+\delta),y)^\top \delta||_2^2.
\end{equation}
By substituting the result from Equation \ref{eq_99} into the above expression, we can obtain the bound presented in Theorem \ref{thm1}.
The bound of $\mathcal{L}(F'(x+\delta),y)$ can similarly be obtained using basic norm inequalities, which we will not elaborate on here.

\textbf{The applicability for big $\delta$.}
In our proof, we apply Taylor expansion which demonstrates effectiveness for small changes in the input, i.e., small $\delta$.
Our discussion extends to the applicability of our theory when faced with sizable $\delta$ values.
In fact, the generation of adversarial examples is an iterative process.
Taylor expansion remains accurate with small update steps.
Moreover, our proof mainly relies on $p(x) \geq p(x + \delta)$.
This implies that as long as we acknowledge that a larger $\delta$ leads to a decrease in $p(x+\delta)$, our theory continues to hold.
It naturally follows that a larger $\delta$ typically correlates with diminished classification accuracy.

\section{Supplementary Experiments}

\subsection{Attack Results on Other Defenses}
\label{eval_other_defenses}

We here evaluate the attack performance of \sysname against other defenses including R\&P \citep{RANDP}, NIPS-R3 \footnote{\url{https://github.com/anlthms/nips-2017/blob/master/poster/defense.pdf}}, FD \citep{FD}, ComDefend \citep{comdefend}, RS \citep{RS}, NRP \citep{NRP}.
These defense mechanisms deviate significantly from adversarial training as they rely on data processing techniques to purify adversarial examples into normal ones.
The detailed settings follow \citet{SSA}.
Table \ref{attack_on_other_defense} reports the attack performance of VT, SSA, RAP, and our method against these defenses.
\sysname still consistently surpasses baselines by a large margin.

\begin{table}[!h]
\caption{The attack success rates (\%) of attacks against various defenses. We use ResNet50 as the proxy model.}
\label{attack_on_other_defense}
\centering
\footnotesize
\begin{tabular}{@{}c|cccccc@{}}
\toprule
Attack & R\&P & NIPS-R3 & FD & ComDefend & RS & NRP \\ \midrule
VT & 64.53 & 70.69 & 73.4 & 87.33 & 52.14 & 43.97 \\
SSA & 72.10 & 75.32 & 76.06 & 91.27 & 61.86 & 52.91 \\
RAP & 93.15 & 92.44 & 92.58 & 95.59 & 76.09 & 76.28 \\
Ours & \textbf{95.81} & \textbf{97.44} & \textbf{94.02} & \textbf{97.64} & \textbf{85.54} & \textbf{83.47} \\ \bottomrule
\end{tabular}
\end{table}

\subsection{Attention Visualization for Targeted Attacks}
\label{appendix_attention}

\begin{figure*}[!h]
    \centering
    \subfloat[Original]{
    \includegraphics[width=0.15\linewidth]{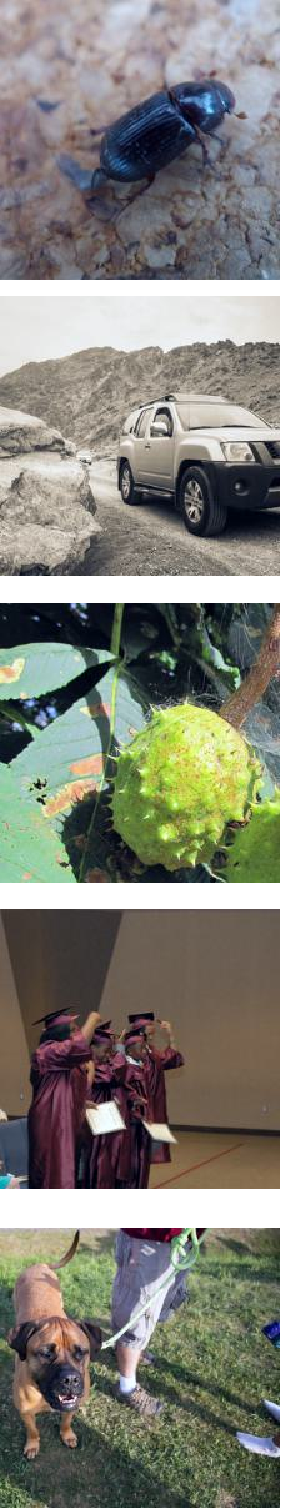}
    }
    \subfloat[VT]{
    \includegraphics[width=0.15\linewidth]{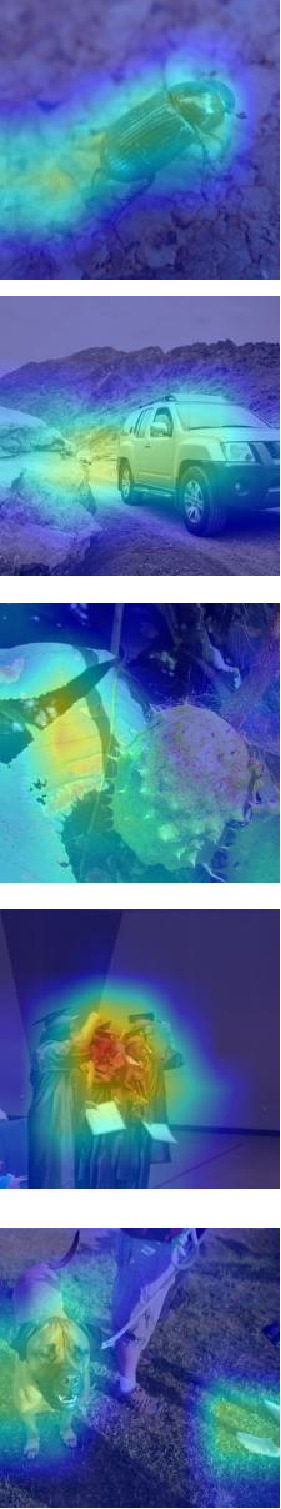}
    }
    \subfloat[SSA]{
    \includegraphics[width=0.15\linewidth]{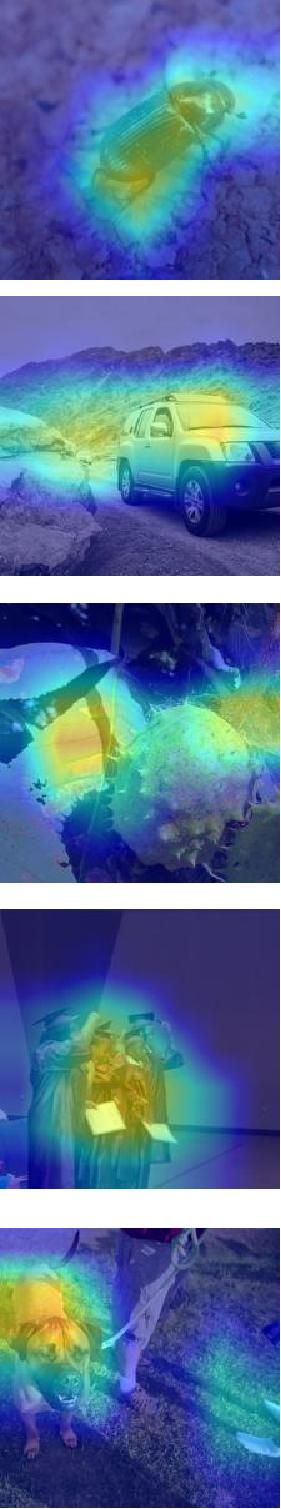}
    }
    \subfloat[Ours]{
    \includegraphics[width=0.15\linewidth]{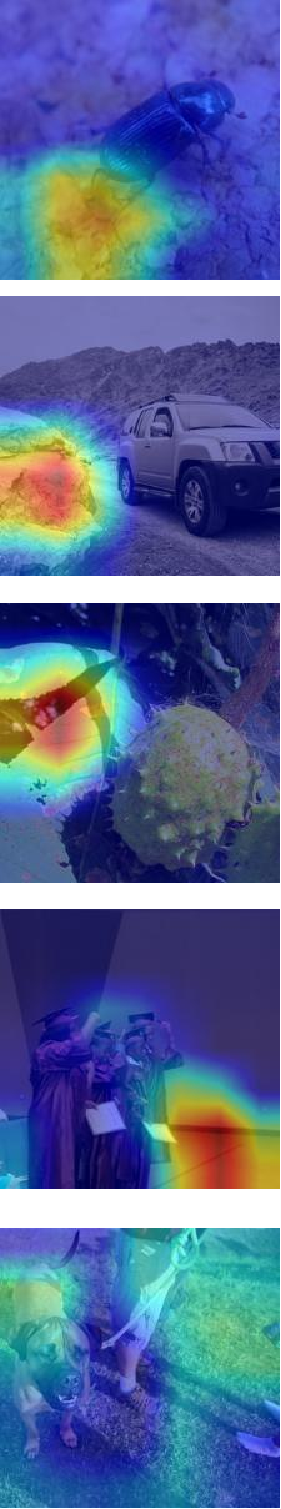}
    }    
    \caption{We conduct targeted attacks and visualize attention maps of the target model to the resultant adversarial images.}
    \label{targeted_attack_visual}
\end{figure*}

Figure \ref{targeted_attack_visual} shows more visualizations of attention maps.
As can be seen, the adversarial examples produced by VT and SSA only can slightly decrease the attention of the target model to entities of images, while \sysname can significantly distract the attention of the target model from the entities to trivial regions.

\subsection{Impact of Hyperparameters}
\label{appendix_sens_impact}

\sysname involves four hyperparameters that can significantly impact the attack performance:

\begin{itemize}
    \item $\lambda$ is a balance factor between the loss of resultant adversarial examples and the flatness of the region around the adversarial ones. The bigger $\lambda$ attaches more attention to promoting adversarial examples toward flat regions.
    \item $b$ suggests how wide the region around adversarial examples is desired to be flat.
    \item When comes to the implementation of \sysname, Hessian matrix is approximately evaluated and a inappropriate $k$ probably induces non-trivial errors.
    \item We extract $N$ samples around adversarial examples to estimate the flatness of the region around the adversarial ones. Generally, the estimation accuracy raises with increasing N.
\end{itemize}

\begin{figure*}[!h]
    \centering
    \subfloat[Impact of $\lambda$]{
    \label{impact_lambda}\includegraphics[width=0.4\linewidth]{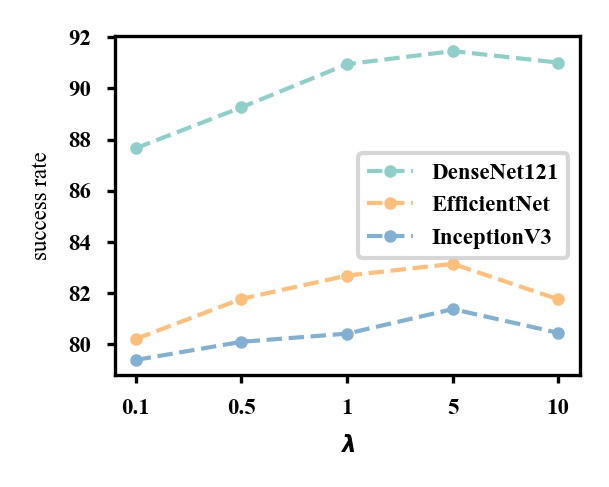}
    }
    \subfloat[Impact of $b$]{
    \label{impact_b}\includegraphics[width=0.4\linewidth]{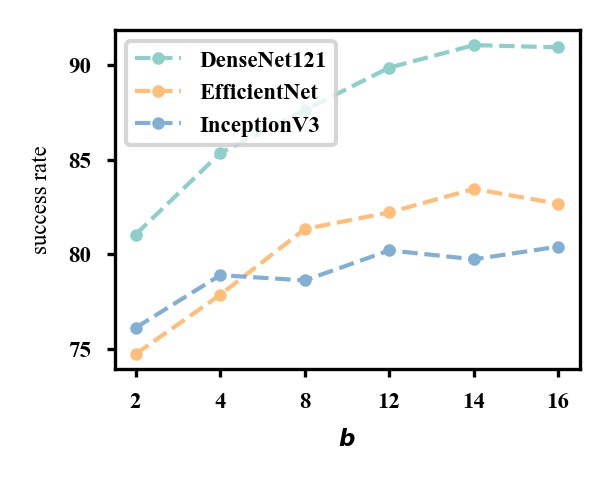}
    }
    
    \subfloat[Impact of $k$]{
    \label{impact_k}\includegraphics[width=0.4\linewidth]{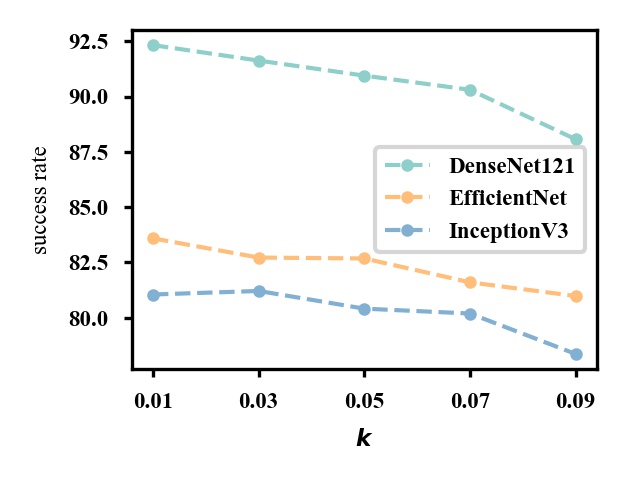}
    }
    \subfloat[Impact of $N$]{
    \label{impact_N}\includegraphics[width=0.4\linewidth]{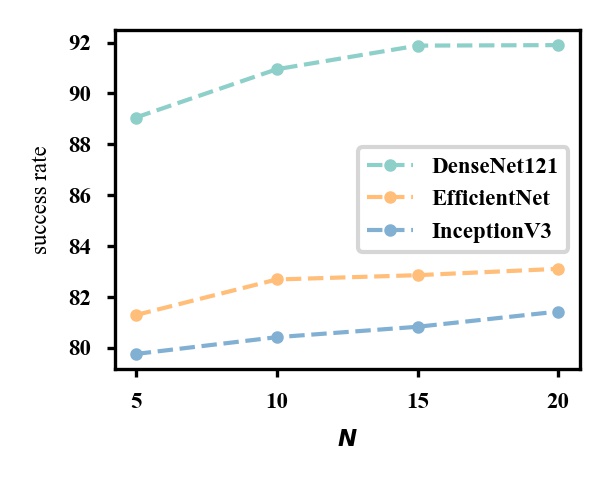}
    }
    \caption{The attack effectiveness of \sysname with varying $\lambda \in \{0.1, 0.5, 1, 5, 10\}, b \in \{1,2,4,8,12,16\},k \in \{0.01, 0.03, 0.05, 0.07, 0.09\},N \in \{5,10,15,20\}$. The proxy model is ResNet50. We set $\epsilon=8$.}
    \label{ablation_study}
\end{figure*}

\textbf{The impact of $\lambda$.}
Figure \ref{ablation_study}(a) illustrates the attack success rates of crafted adversarial examples by \sysname with varying $\lambda$ against target models.
We see that as $\lambda$ increases, the attack effectiveness of \sysname presents the tendency to climb initially and decline thereafter.
If $\lambda$ is small like $\lambda=0.1$, most attention of Equation \ref{our_optim} is put into optimizing $L$ while ignoring the flatness of the region around crafted adversarial examples, \textit{i.e.}, degrading into vanilla transfer-based attacks and causing that the adversarial ones are more likely to be trapped by model-specified sharp regions.
As a remedy, properly increasing $\lambda$ can grab part of the attention of Equation \ref{our_optim} to focus on the flatness of the surrounding region, so that the resultant adversarial examples have more chance to evade the model-specified sharp regions and are more transferable.
However, it should be stressed that too bigger $\lambda$ induces Equation \ref{our_optim} to only focus on the flatness and de-emphasize whether the region poses a threat to the proxy model, leading to a reduction in the attack success rate.
Therefore, carefully adjusting $\lambda$ is necessary.

\textbf{The impact of $b$.}
We examine the attack effectiveness of \sysname by changing the hyperparamter $b$ and Figure \ref{ablation_study}(b) shows the attack results over different proxy-target pairs.
Overall, the attack performance of \sysname steadily increases with increasing $b$.
The reason for it is that, a small $b$ makes the extracted samples too close to the resultant adversarial examples and causes that the flatness of the region around the resultant adversarial examples cannot be effectively evaluated.
Furthermore, increasing $b$ alleviates the issue so as to boost the transferability of produced adversarial examples.
Moreover, interestingly, we find that, when employing ResNet50 and EfficientNet, increasing $b$ from 14 to 16 slightly hurts the transferability of produced adversarial examples, and this is probably attributed to the extracted samples being a little far from the crafted adversarial examples.

\textbf{The impact of $k$.}
Figure \ref{ablation_study}(c) shows the influence of the hyperparameter $k$ to the performance of \sysname and we observe that increasing $k$ weakens the attack effectiveness of \sysname.
In fact, as shown in Section \ref{approach_approximate_solution}, Hessian matrix is approximately replaced by gradient difference, \textit{i.e.}, Equation \ref{appro_res}, and the feasibility of the approximation solution depends on a small $k$ to omit the error of approximation induced by Taylor expansion.
Hence, it is intuitive why a bigger $k$ incurs degradation on the attack success rates of \sysname.

\textbf{The impact of $N$.}
Figure \ref{ablation_study}(d) shows the attack performance of \sysname with different $N$.
Intuitively, by sampling more instances from the region around the resultant adversarial examples, \textit{i.e.}, increasing $N$, we can make a more accurate estimation of the flatness of the region, which in turn can decrease estimation errors and then strengthen the transferability of generated adversarial examples.
As expected, Figure \ref{ablation_study}(d) validates this point that increasing N promotes the transferability of produced adversarial examples.

\subsection{Attacks with Varying Iteration}
\label{appendix_varying_iter}

We here investigate the impact of iterations on attack performance.
Table \ref{iteration_impact} reports the performance of different attacks with varying numbers of iterations.
It can be observed that there is negligable change in attack performance when the number of iterations is increased from 10 to 20.
The results suggest that the attack methods achieve convergence with 10 iterations.

\begin{table}[!h]
\caption{The attack effectiveness of different attacks with varying iterations. We use ResNet50 as the proxy model.}
\label{iteration_impact}
\centering
\begin{tabular}{@{}cccc@{}}
\toprule
Attack & ResNet50 & DenseNet121 & EfficientNet \\ \midrule
VT (10 iter) & 100.00 & 88.76 & 81.51 \\
VT (20 iter) & 100.00 & 89.62 & 81.63 \\
SSA (10 iter) & 100.00 & 95.29 & 90.72 \\
SSA (20 iter) & 100.00 & 95.51 & 91.56 \\
Ours (10 iter) & 99.80 & 99.08 & 99.25 \\
Ours (20 iter) & 99.85 & 99.69 & 99.52 \\ \bottomrule
\end{tabular}
\end{table}

\section{Score Implication}
\label{appendix_score_implication}
We begin by gathering predictions from Google MLaaS platforms for both label detection and object detection over crafted adversarial examples.
We assign scores to these predictions along five discrete levels, with scores of 1 through 5 indicating the degree of accuracy: totally wrong, slightly incorrect, strange but not incorrect, relatively reasonable, and precise.
A score of 1 suggests that the images do not contain the objects predicted.
A score of 2 indicates minor errors in the predictions, such as identifying flowers instead of trees in a tree image.
A score of 3 suggests that the main object in the image is not correctly recognized, for example, predicting stones, roads, or tires for a car image.
A score of 4 denotes accurately identifying the general type of the main object in the images, while a score of 5 indicates that the system can fully and correctly identify the main object in the image.
Next, a volunteer (See Section \ref{ethic_statement} for more information on the recruitment of volunteers and the evaluation procedure) is asked to rate the consistency between the predictions and the images based on these scores
For search engines, a similar evaluation procedure is followed, where we assess the similarity between the retrieved images and their corresponding original images.

\section{A visualization of different attacks against Google Service and Search Engines}
\label{appendix_visual_real_world}

\begin{figure*}[!h]

    \centering
    
    \subfloat[Original-Classification]{
    \includegraphics[width=0.4\linewidth]{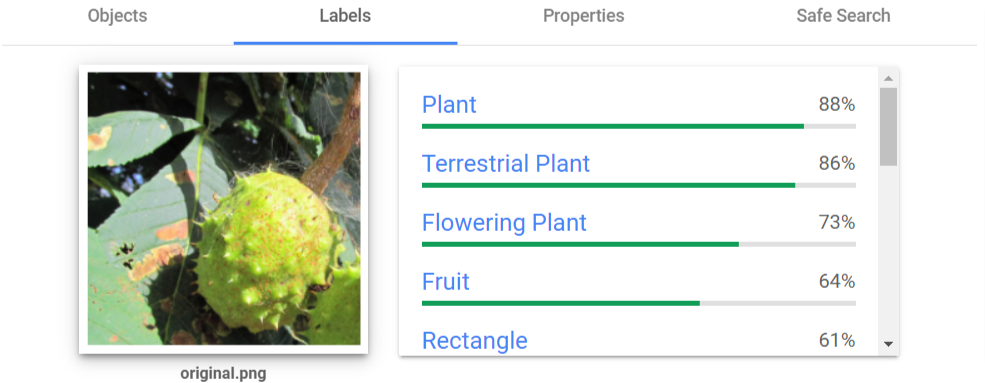}
    }
    \subfloat[VT-Classification]{
    \includegraphics[width=0.4\linewidth]{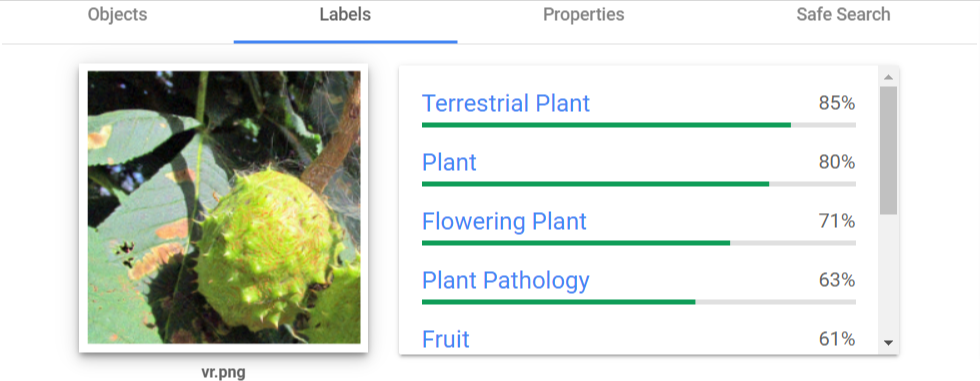}
    }
    
    \subfloat[SSA-Classification]{
    \includegraphics[width=0.4\linewidth]{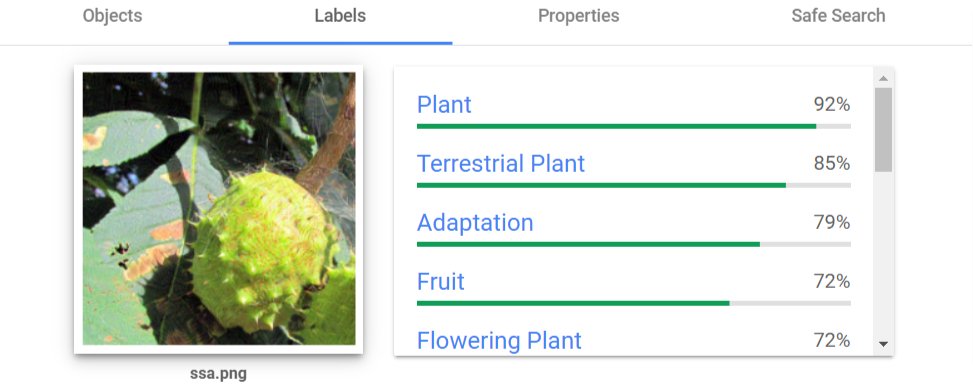}
    }
    \subfloat[Ours-Classification]{
    \includegraphics[width=0.4\linewidth]{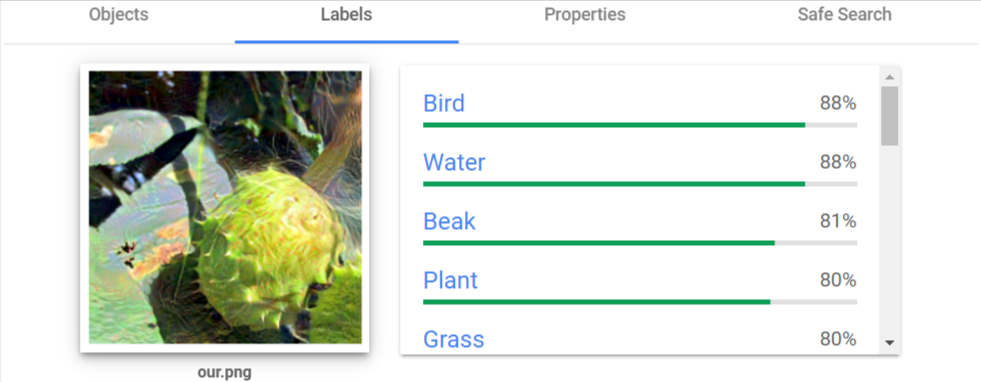}
    }
    
    \subfloat[Original-Object-Detection]{
    \includegraphics[width=0.4\linewidth]{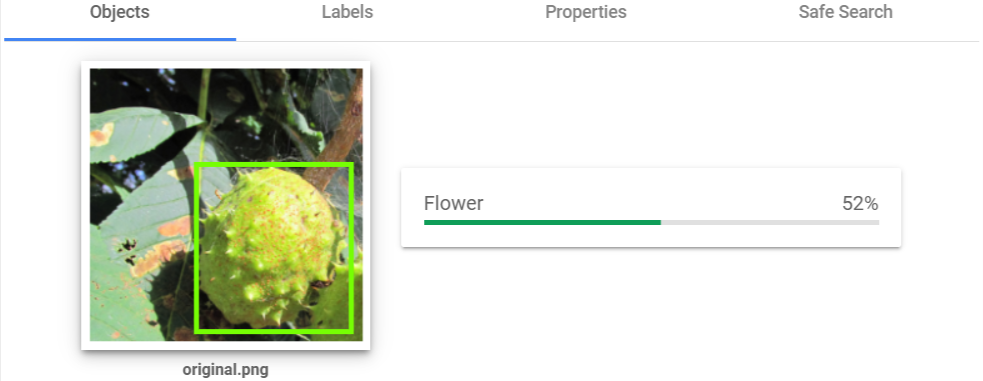}
    }
    \subfloat[VT-Object-Detection]{
    \includegraphics[width=0.4\linewidth]{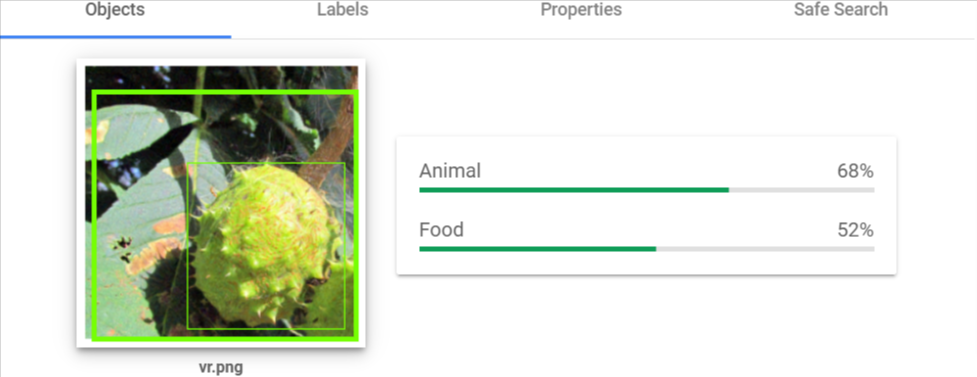}
    }
    
    \subfloat[SSA-Object-Detection]{
    \includegraphics[width=0.4\linewidth]{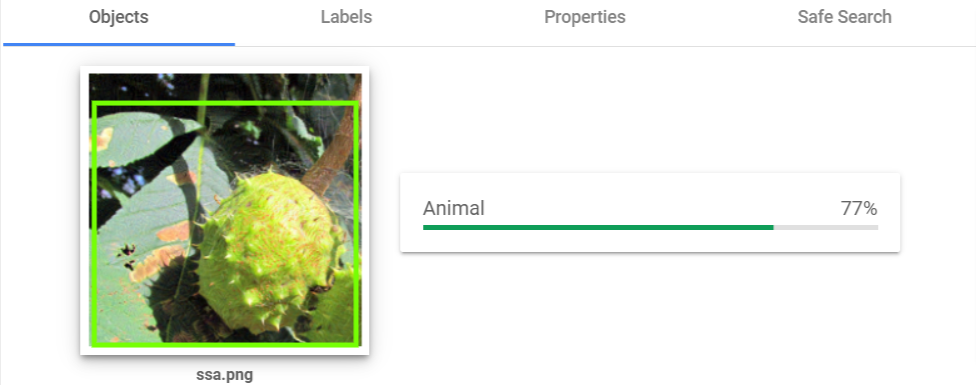}
    }
    \subfloat[Ours-Object-Detection]{
    \includegraphics[width=0.4\linewidth]{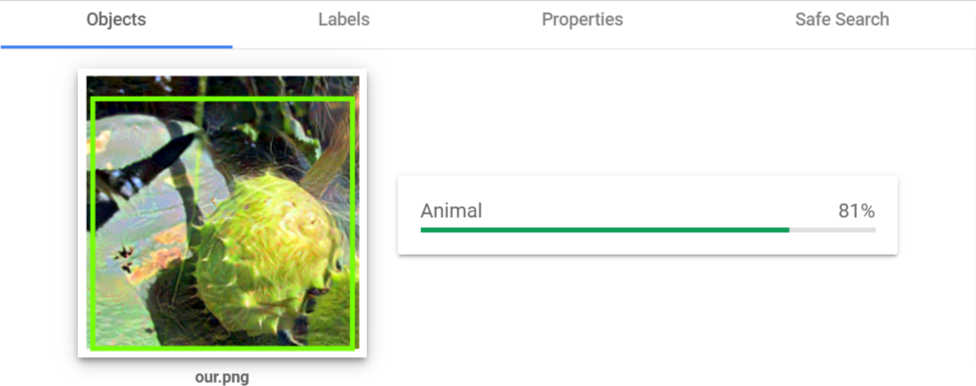}
    }

    
    
    
    
    
    
    
    \caption{The attack results on Google Cloud Vision including image classification and object detection tasks. We do not know any knowledge of it and the proxy model is ResNet50.}
    \label{google_test}
\end{figure*}

Figure \ref{google_test} shows a visualization of different attacks against Google Service, and we use the format \{attack method\}-\{task\} to denote the attack used to craft adversarial examples and the corresponding test task.

For image classification, consistent with our intuition, the original image is predicted into plant-related categories. However, the returned predictions for the adversarial examples produced via VT and SSA are close to the ground-truth label of the original image. Therefore, the adversarial examples cannot be deemed to be threats against Google Cloud Vision. In contrast, the adversarial examples generated by our method indeed mislead Google Cloud Vision, where the predictions with the highest confidence are bird, water, and beak, and these predictions are fairly irrelevant to the original images.

For object detection, we can obtain similar conclusions. The original image is correctly detected. The adversarial examples produced by VT seem to fool Google Cloud Vision to some extent, while SSA can fully mislead Google Cloud Vision. Also, our adversarial examples trick Google Cloud Vision with higher confidence than VT.

\begin{figure*}[!th]
    \centering

    \subfloat[Original-Google]{
    \includegraphics[width=0.4\linewidth]{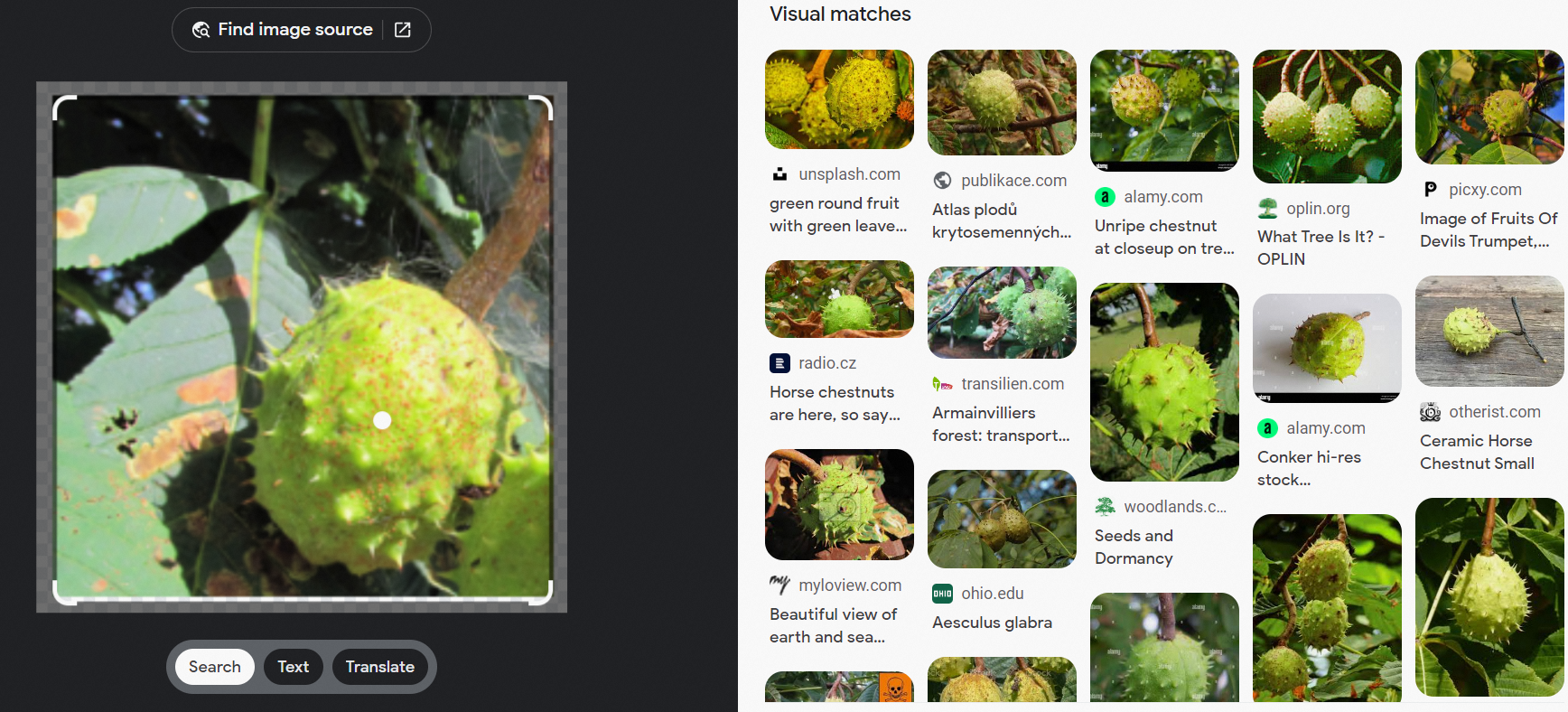}
    }
    \subfloat[Ours-Google]{
    \includegraphics[width=0.4\linewidth]{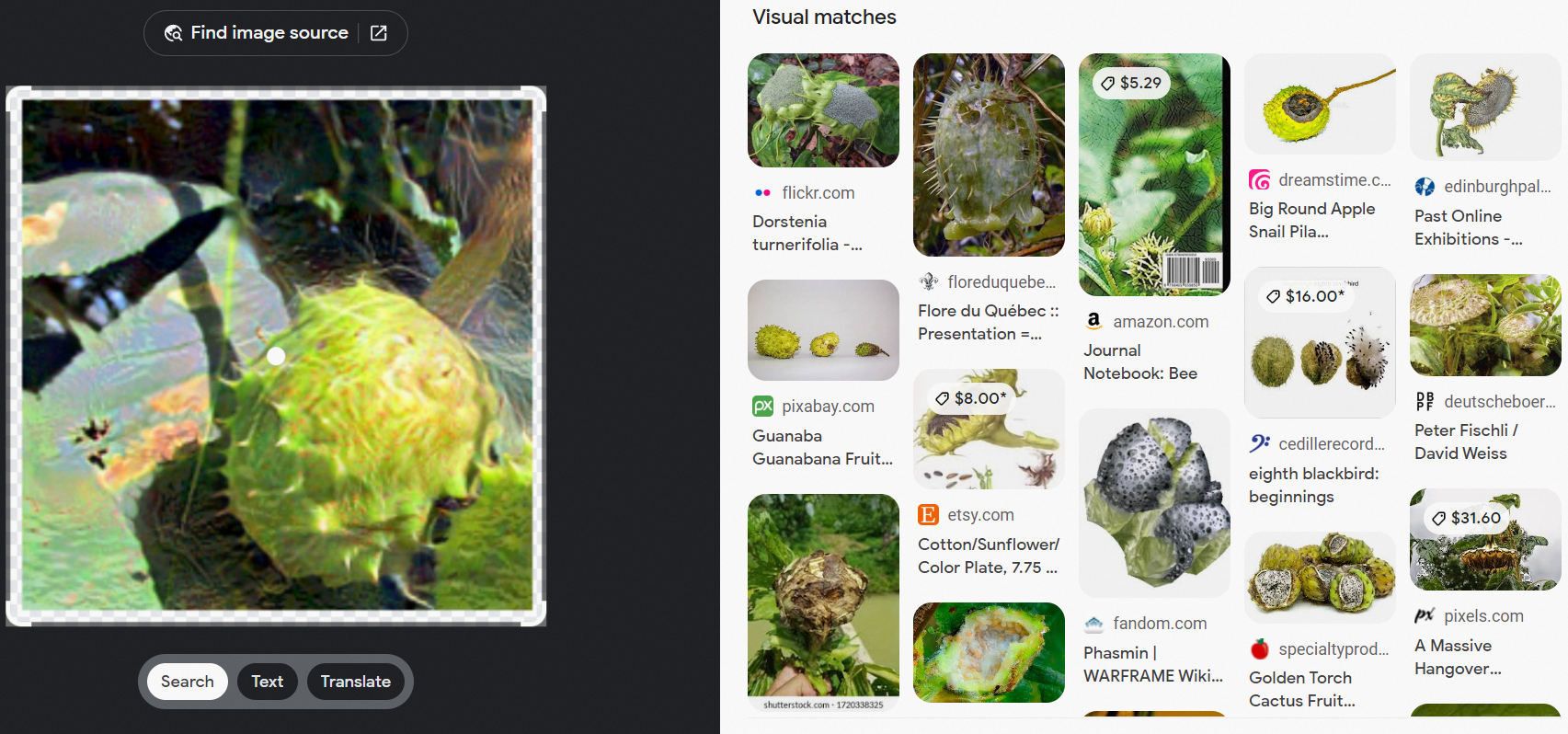}
    }
    
    \subfloat[Original-Bing]{
    \includegraphics[width=0.4\linewidth]{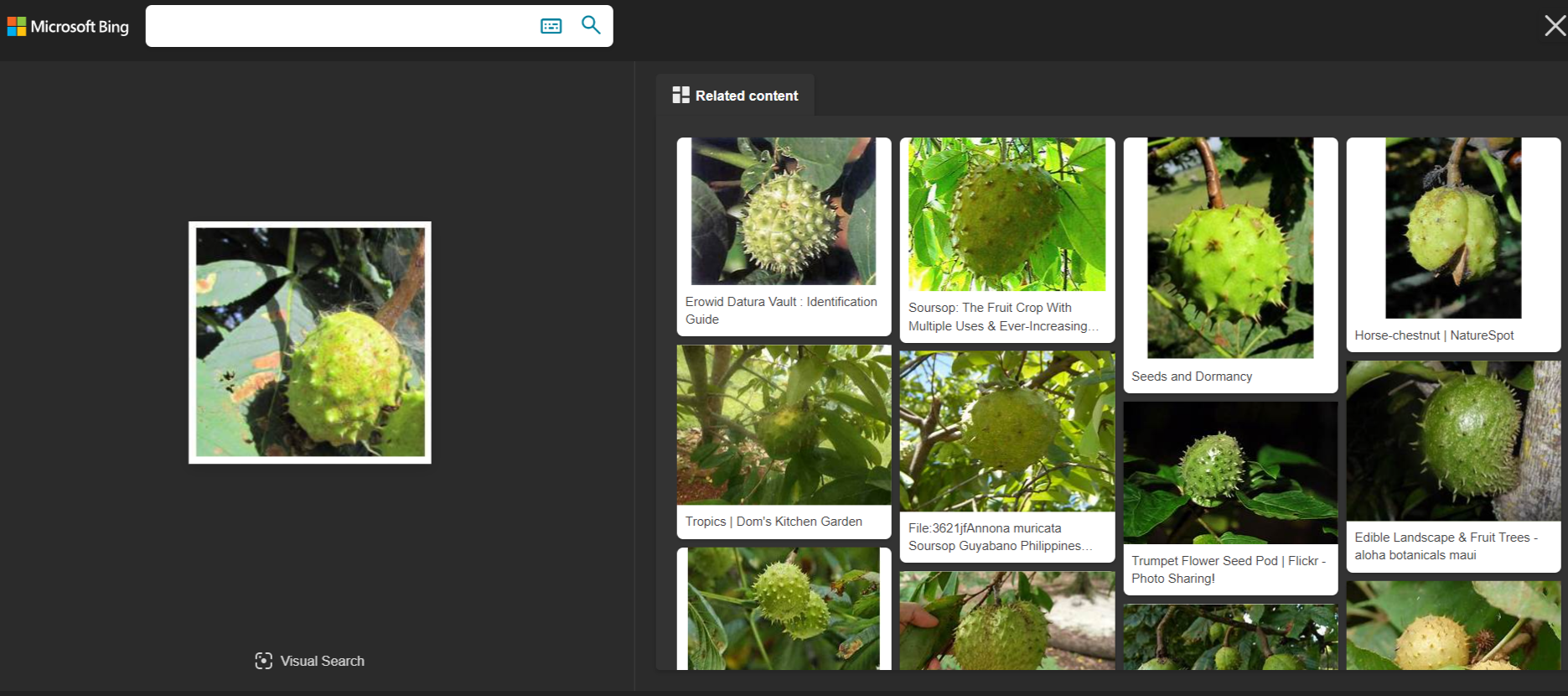}
    }
    \subfloat[Ours-Bing]{
    \includegraphics[width=0.4\linewidth]{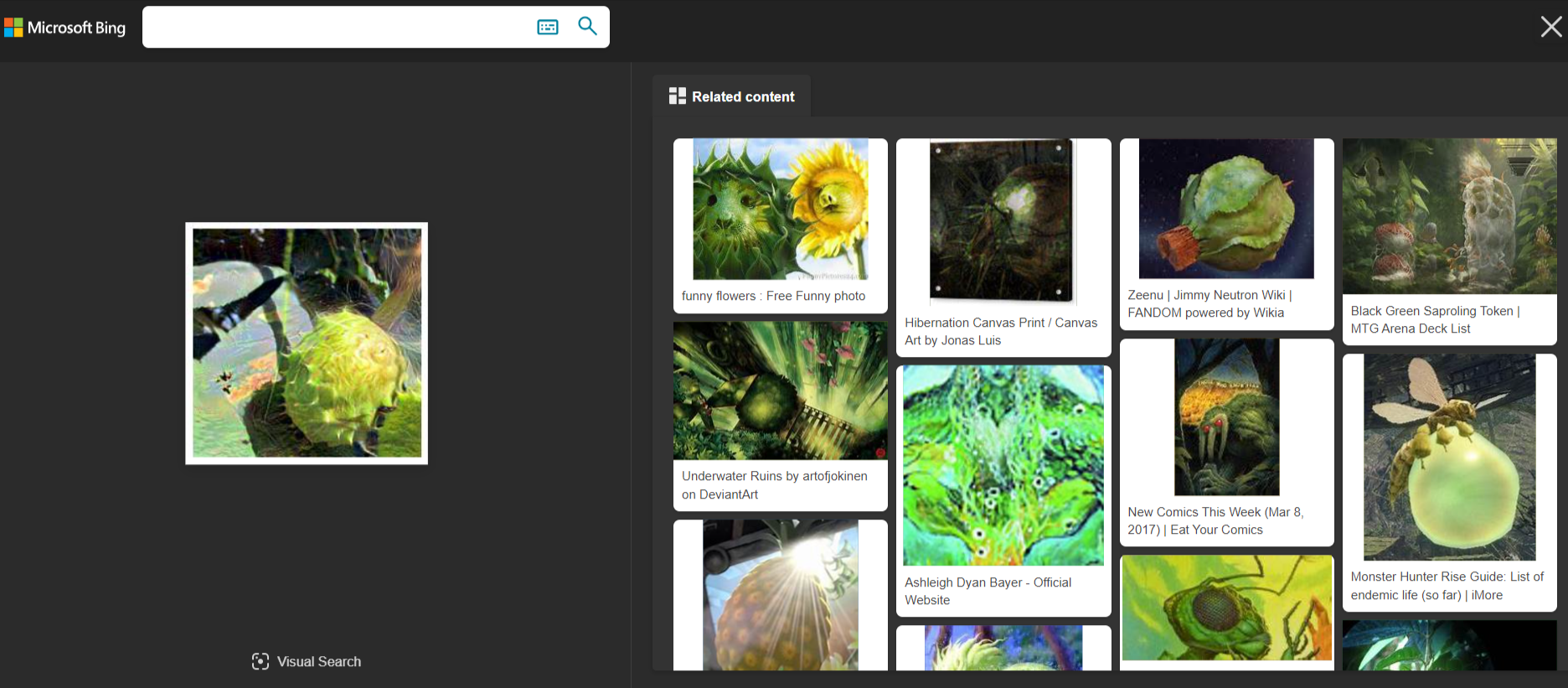}
    }
    
    \subfloat[Original-Yandex]{
    \includegraphics[width=0.4\linewidth]{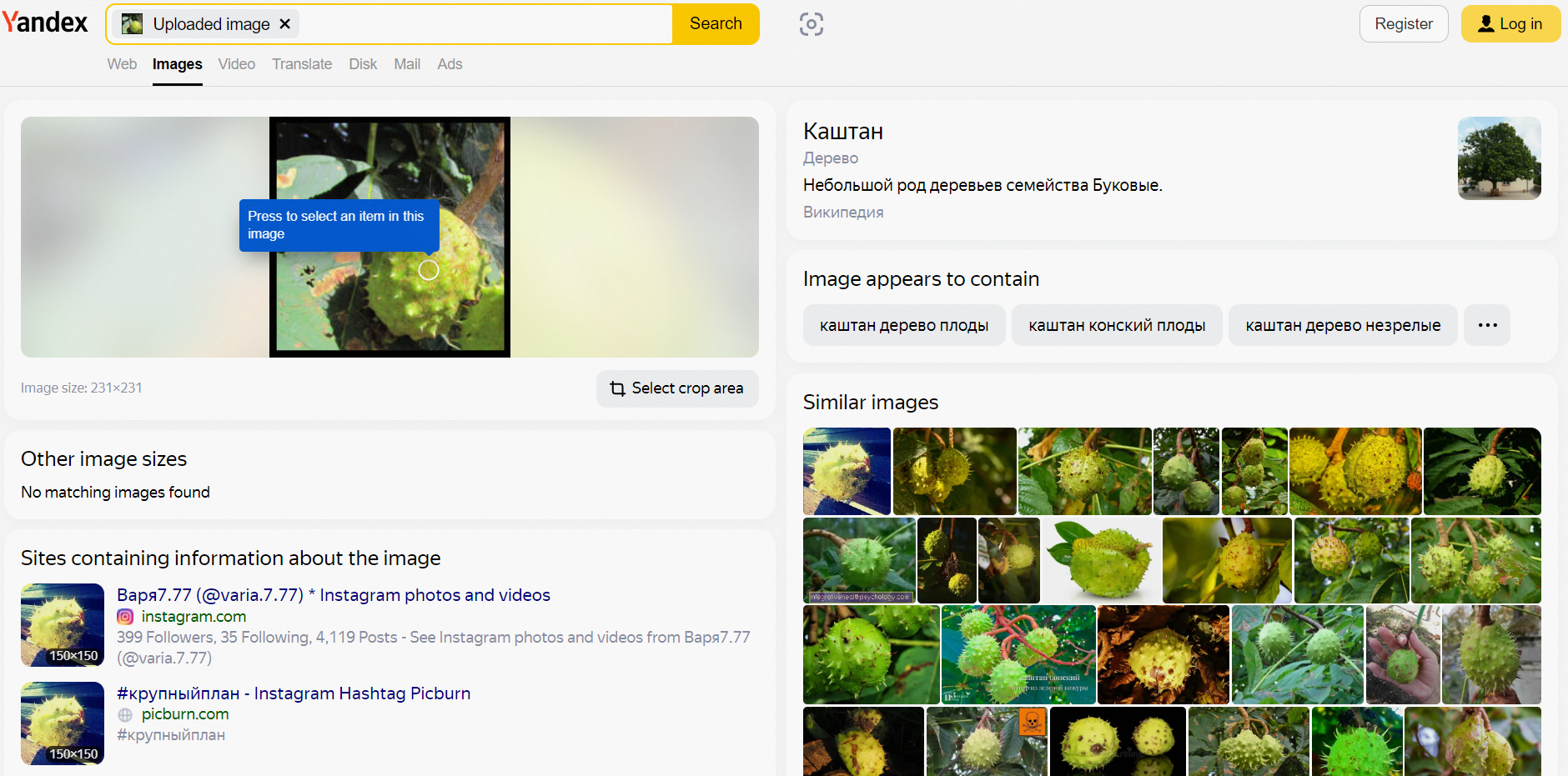}
    }
    \subfloat[Ours-Yandex]{
    \includegraphics[width=0.4\linewidth]{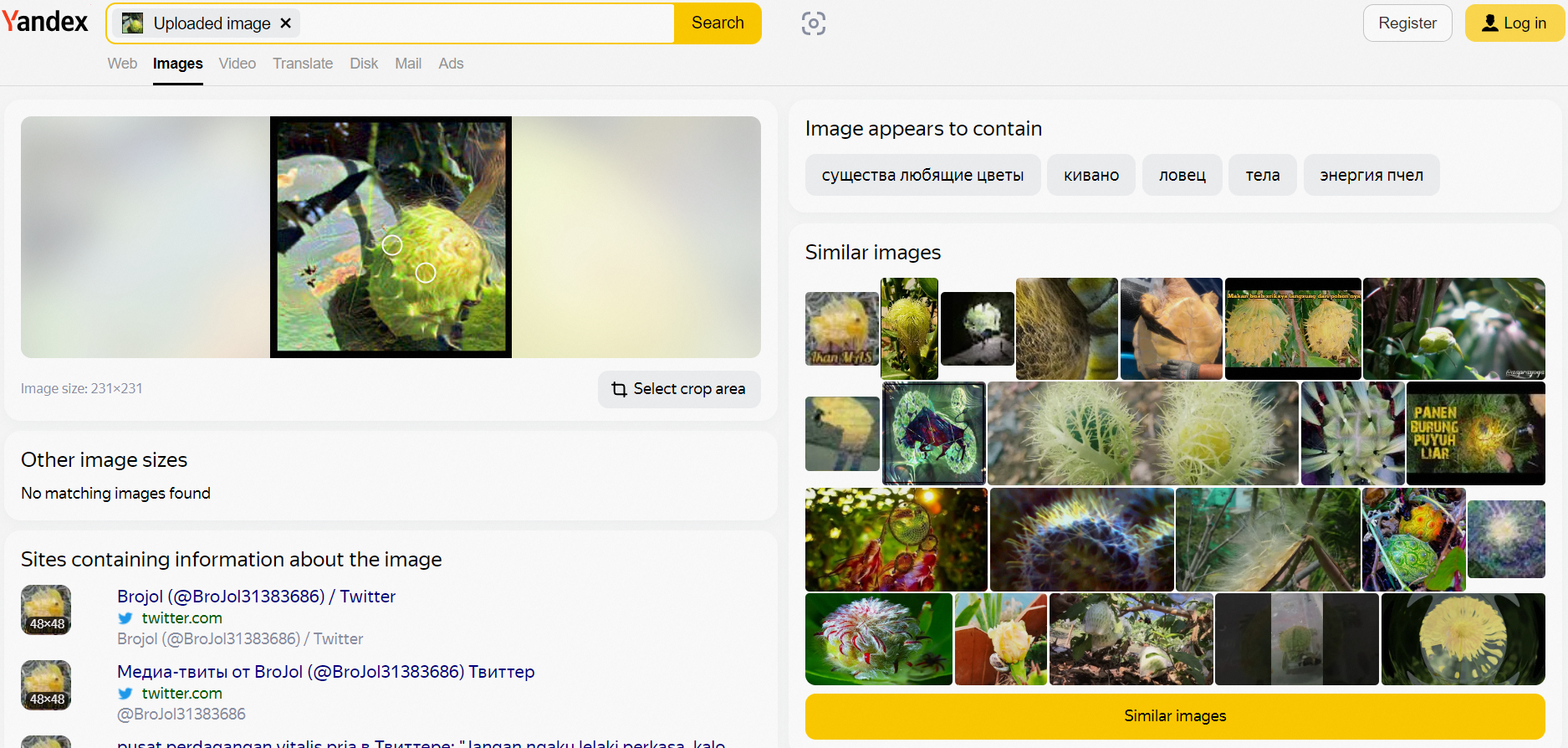}
    }
    
    \subfloat[Original-Baidu]{
    \includegraphics[width=0.4\linewidth]{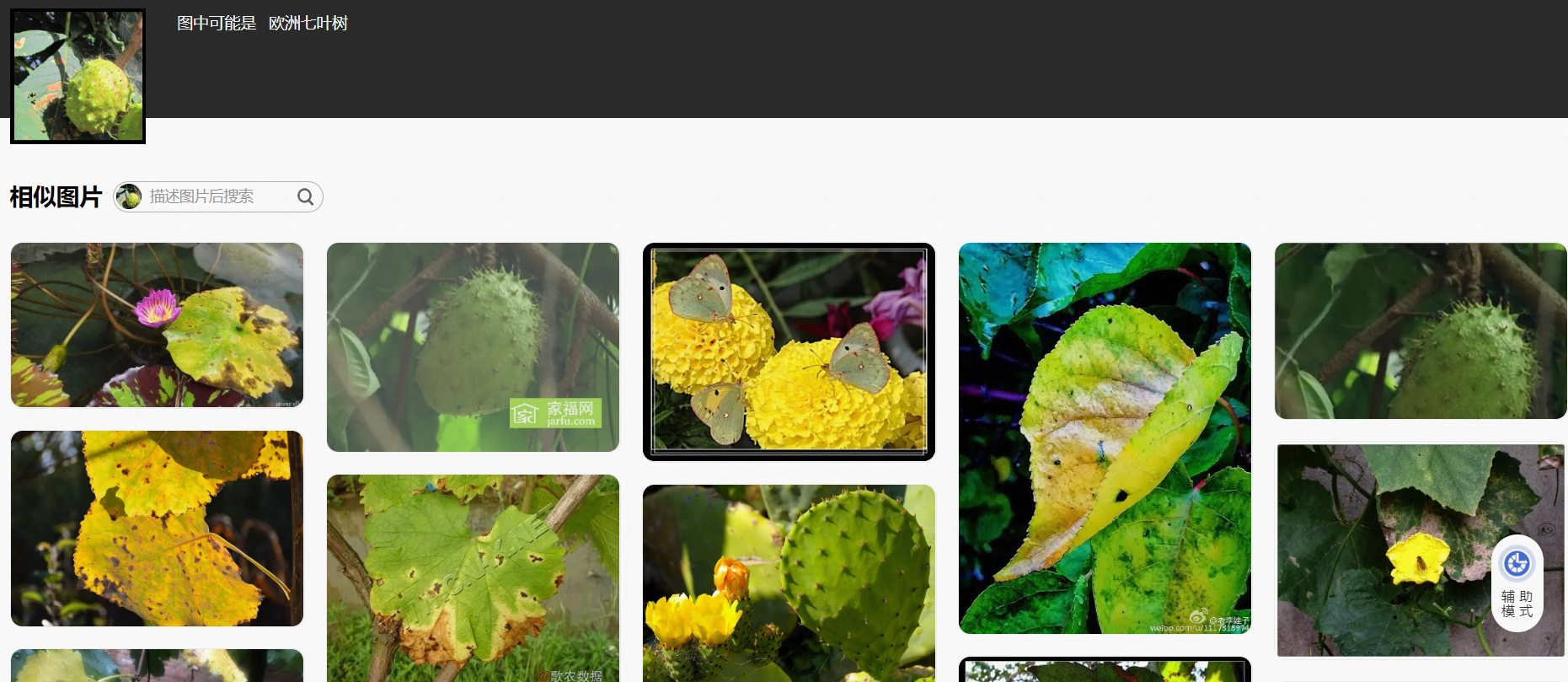}
    }
    \subfloat[Ours-Baidu]{
    \includegraphics[width=0.4\linewidth]{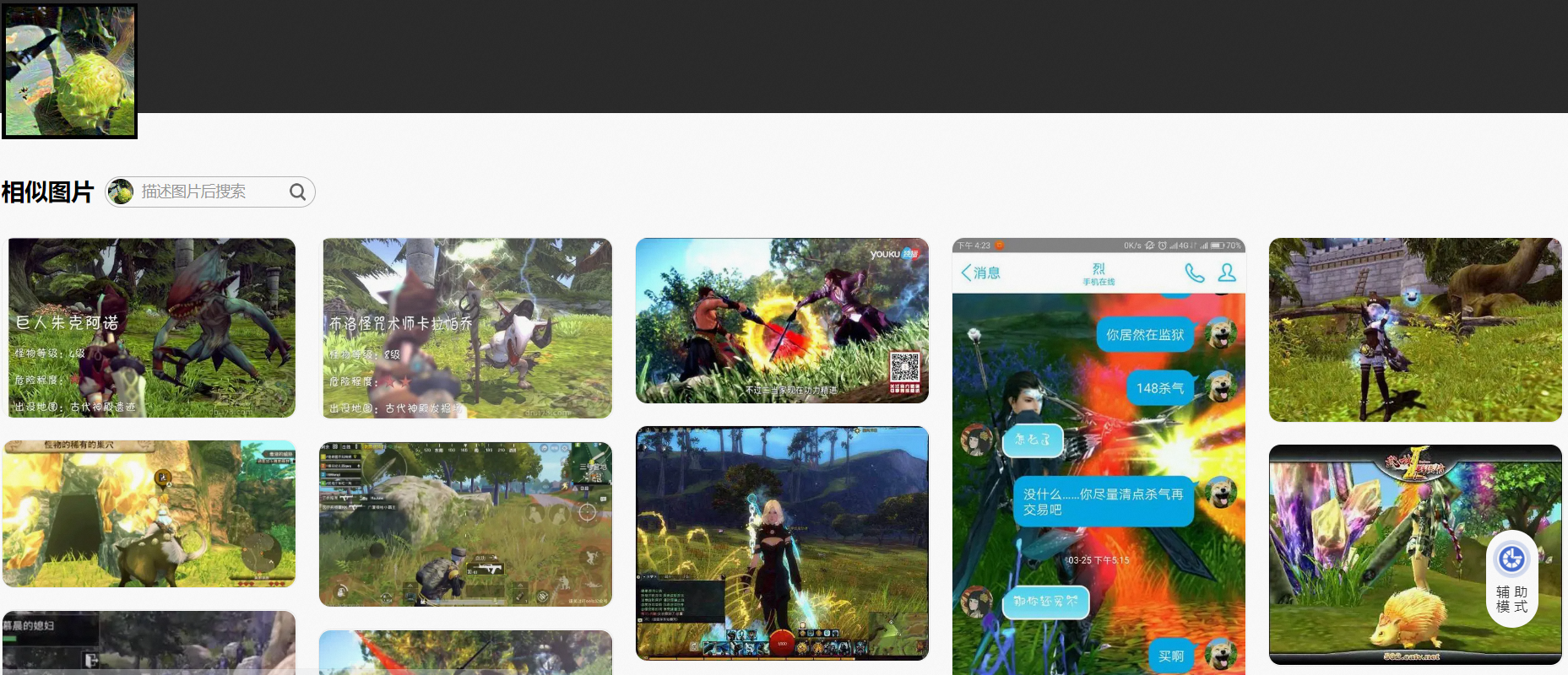}
    }

    
    
    
    
    
    
    
    \caption{An example for attacking four state-of-the-art search engines.}
    \label{search_engine_visual}
\end{figure*}

Figure \ref{search_engine_visual} visualizes an example of \sysname against four state-of-the-art search engines.
We observe that search engines fetch high-quality and similar images for normal samples.
However, when we input the generated adversarial examples, the quality of retrieved images noticeably deteriorates, particularly in the case of Baidu.

\section{Ethics Statement}
\label{ethic_statement}

This paper designs a novel approach to enhance the transferability of adversarial examples.
While this approach is easy-to-implement and seems harmful, it is believed that the benefits of publishing \sysname outweigh the potential harms.
It is better to expose the blind spots of DNNs as soon as feasible because doing so can alert deployers to be aware of potential threats and greatly encourage AI community to design corresponding defense strategies.

For the human assessment process (Section \ref{real_eval}), we generated adversarial examples and collected the predictions of different applications on these adversarial examples.
Throughout the entire process, all communication with the volunteer, including recruitment, was conducted anonymously.
Similarly, we do not knew the personal information regarding to the volunteer.
The volunteer were unaware of our specific objectives, ensuring that there was no interest between the volunteer and us.
The volunteer came from a certain university and possessed normal discernment abilities.
During the rating process, the volunteer were unaware of whether a given sample was an adversarial example or the attack method used to generate it.
Therefore, there was no bias towards any particular type of attack method from the volunteer.
Overall, the evaluation process was relatively fair.

\newpage
\section*{NeurIPS Paper Checklist}


\begin{enumerate}

\item {\bf Claims}
    \item[] Question: Do the main claims made in the abstract and introduction accurately reflect the paper's contributions and scope?
    \item[] Answer: \answerYes{} 
    \item[] Justification: The main claims reflect the contributions.

\item {\bf Limitations}
    \item[] Question: Does the paper discuss the limitations of the work performed by the authors?
    \item[] Answer: \answerYes{} 
    \item[] Justification: Our appendix contains ethics statement.

\item {\bf Theory Assumptions and Proofs}
    \item[] Question: For each theoretical result, does the paper provide the full set of assumptions and a complete (and correct) proof?
    \item[] Answer: \answerYes{} 
    \item[] Justification: Our appendix contains complete proofs.

    \item {\bf Experimental Result Reproducibility}
    \item[] Question: Does the paper fully disclose all the information needed to reproduce the main experimental results of the paper to the extent that it affects the main claims and/or conclusions of the paper (regardless of whether the code and data are provided or not)?
    \item[] Answer: \answerYes{} 
    \item[] Justification: We provide source codes.

\item {\bf Open access to data and code}
    \item[] Question: Does the paper provide open access to the data and code, with sufficient instructions to faithfully reproduce the main experimental results, as described in supplemental material?
    \item[] Answer: \answerYes{} 
    \item[] Justification: We provide source codes.

\item {\bf Experimental Setting/Details}
    \item[] Question: Does the paper specify all the training and test details (e.g., data splits, hyperparameters, how they were chosen, type of optimizer, etc.) necessary to understand the results?
    \item[] Answer: \answerYes{} 
    \item[] Justification: We make detailed experiment settings in our paper.

\item {\bf Experiment Statistical Significance}
    \item[] Question: Does the paper report error bars suitably and correctly defined or other appropriate information about the statistical significance of the experiments?
    \item[] Answer: \answerNo{} 
    \item[] Justification: We only run one trial.

\item {\bf Experiments Compute Resources}
    \item[] Question: For each experiment, does the paper provide sufficient information on the computer resources (type of compute workers, memory, time of execution) needed to reproduce the experiments?
    \item[] Answer: \answerNo{} 
    \item[] Justification: Common GPUs are capable of running our experiments.
    
\item {\bf Code Of Ethics}
    \item[] Question: Does the research conducted in the paper conform, in every respect, with the NeurIPS Code of Ethics \url{https://neurips.cc/public/EthicsGuidelines}?
    \item[] Answer: \answerYes{} 
    \item[] Justification: We follow code of ethics.

\item {\bf Broader Impacts}
    \item[] Question: Does the paper discuss both potential positive societal impacts and negative societal impacts of the work performed?
    \item[] Answer: \answerYes{} 
    \item[] Justification: We explain this in Introduction.
    
\item {\bf Safeguards}
    \item[] Question: Does the paper describe safeguards that have been put in place for responsible release of data or models that have a high risk for misuse (e.g., pretrained language models, image generators, or scraped datasets)?
    \item[] Answer: \answerNA{} 
    \item[] Justification: This paper does not releases any new models or datasets.

\item {\bf Licenses for existing assets}
    \item[] Question: Are the creators or original owners of assets (e.g., code, data, models), used in the paper, properly credited and are the license and terms of use explicitly mentioned and properly respected?
    \item[] Answer: \answerYes{} 
    \item[] Justification: We do this.

\item {\bf New Assets}
    \item[] Question: Are new assets introduced in the paper well documented and is the documentation provided alongside the assets?
    \item[] Answer: \answerNA{} 
    \item[] Justification: This paper does not release new assets.

\item {\bf Crowdsourcing and Research with Human Subjects}
    \item[] Question: For crowdsourcing experiments and research with human subjects, does the paper include the full text of instructions given to participants and screenshots, if applicable, as well as details about compensation (if any)? 
    \item[] Answer: \answerYes{} 
    \item[] Justification: Our appendix contains this.

\item {\bf Institutional Review Board (IRB) Approvals or Equivalent for Research with Human Subjects}
    \item[] Question: Does the paper describe potential risks incurred by study participants, whether such risks were disclosed to the subjects, and whether Institutional Review Board (IRB) approvals (or an equivalent approval/review based on the requirements of your country or institution) were obtained?
    \item[] Answer: \answerNA{} 
    \item[] Justification: There are no risks in our experiment for study participants.

\end{enumerate}

\end{document}